\documentclass[10pt,a4paper,american]{amsart}
\usepackage[left=3.5cm, right=3.5cm]{geometry}
\usepackage[american]{babel}
\usepackage{tikz}
\usetikzlibrary{chains, positioning}
\usepackage{amsthm}
\usepackage{amsbsy}
\usepackage{amstext}
\usepackage{amsfonts}
\usepackage{mathcomp}
\usepackage{dsfont}
\usepackage{latexsym}
\usepackage{amssymb}
\usepackage{esint}
\usepackage[colorlinks,pdfpagelabels,pdfstartview = FitH,bookmarksopen = true,bookmarksnumbered = true,linkcolor = blue,plainpages = false,hypertexnames = false,citecolor = red,pagebackref=true]{hyperref}
\usepackage{cite}
\usepackage{stmaryrd}
\usepackage{comment}
\usepackage{graphicx}
\usepackage{calc}
\usepackage{flafter}

\newtheorem{thm}{Theorem}[section]
\newtheorem{lem}[thm]{Lemma}
\newtheorem{prop}[thm]{Proposition}

\theoremstyle{definition}
\newtheorem{defn}[thm]{Definition}

\newcommand{\norm}[1]{\lVert#1\rVert}

\DeclareMathOperator*{\diam}{diam}
\DeclareMathOperator*{\Lip}{Lip}
\DeclareMathOperator*{\tr}{tr}
\DeclareMathOperator*{\adj}{adj}
\DeclareMathOperator*{\perm}{perm}

\def\XXint#1#2#3{{\setbox0=\hbox{$#1{#2#3}{\int}$ }
\vcenter{\hbox{$#2#3$ }}\kern-.6\wd0}}

\def\XXiint#1#2#3{{\setbox0=\hbox{$#1{#2#3}{\iint}$ }
\vcenter{\hbox{$#2#3$ }}\kern-.55\wd0}}

\renewcommand{\d}{\:\:\!\!\mathrm{d}}

\newcommand{\R}{\ensuremath{\mathbb{R}}}

\numberwithin{equation}{section}
\allowdisplaybreaks

\begin{document}
\renewcommand{\refname}{References}
\renewcommand{\abstractname}{Abstract}

\title[Solving the Monge-Amp\`ere equation using neural networks]{Solving the Dirichlet problem for the Monge-Amp\`ere equation using neural networks}
\date{\today}

\makeatother
\thanks{K.N was partially supported by grant  2017-03805 from the Swedish research council (VR)}

\author[K. Nystr\"om]{Kaj Nystr\"om}
\address{Kaj Nystr\"om \\
Department of Mathematics, Uppsala University\\
S-751 06 Uppsala, Sweden}
\email{kaj.nystrom@math.uu.se}

\author[M. Vestberg]{Matias Vestberg}
\address{Matias Vestberg\\
Department of Mathematics, Uppsala University\\
S-751 06 Uppsala, Sweden}
\email{matias.vestberg@math.uu.se}

\begin{abstract}
The  Monge-Amp\`ere equation is a fully nonlinear partial differential equation (PDE) of fundamental importance in analysis, geometry and in the applied sciences. In this paper we solve the Dirichlet problem associated with the Monge-Amp\`ere equation using neural networks and we show that an ansatz using deep input convex neural networks can be used to find the unique convex solution. As part of our analysis we study the effect of singularities, discontinuities and noise in the source function, we consider nontrivial domains, and we investigate how the method performs in higher dimensions. We investigate the convergence numerically and present error estimates based on a stability result. We also compare this method to an alternative approach in which standard feed-forward networks are used together with a loss function which penalizes lack of convexity.
\end{abstract}
\maketitle

\section{Introduction}
The Monge-Amp\`{e}re equation
$$\det D^2u = f(x,\,u,\,\nabla u),$$
for an unknown convex function $u:\mathbb{R}^n\to\mathbb R$, $n\geq 1$, where $D^2 u$ is the Hessian of $u$ and $\det D^2 u$ denotes the determinant of $ D^2 u$,
appears naturally in many geometric applications such as the prescribed Gaussian curvature problem, problems in affine geometry \cite{Ca, CheY, Po, TruWa2}, and in physical applications such as inverse reflection and refraction \cite{BriHaPla, BriHaPla2, KaWa}. For example, the Gaussian curvature $K(x)$ of the graph of a function $u:\mathbb{R}^n\to\mathbb R$ at $(x,\,u(x))$ is given by
$$\det D^2u = K(x)(1 + |\nabla u|^2)^{\frac{n+2}{2}}.$$
Another important application is the Monge problem in transport theory, which for a range of cost functions can be expressed using Monge-Amp\`ere type equations coupled with a suitable boundary condition \cite{TruWa, Ur}.  In the case of a quadratic cost function and given probability densities $f_1,\,f_2$ supported on domains $\Omega_1,\, \Omega_2\subset\mathbb{R}^n$, the optimal transport problem is to minimize the transport cost
$$J(T) = \int_{\Omega_1} |T(x) - x|^2f_1(x)\,\d x,$$
over measure-preserving maps $T: \Omega_1 \rightarrow \Omega_2$, i.e. over all maps $T$ such that $$f_1(x)\,\d x = f_2(T(x))\det DT(x)\, \d x.$$ A theorem of Brenier \cite{Br} states that the optimal map exists and that it is given as the gradient of a convex function $u$ on $\Omega_1$. The measure-preserving condition implies that $u$ has to satisfy the Monge-Amp\`{e}re equation
$$\det D^2u = \frac{f_1(x)}{f_2(\nabla u(x))},$$
in a sense that has to be made precise.  In some of the applications mentioned the Monge-Amp\`ere equation is generalized further by adding a matrix possibly depending on $x$, $u$ and $\nabla u$ to the Hessian, before taking the determinant.

In this paper we give a numerical contribution  to the study of the Dirichlet problem for the Monge-Amp\`ere equation. More specifically we use an ansatz based on (deep) neural networks to solve
\begin{align}\label{prob:MA_D}
\left\{
\begin{array}{ll}
\det D^2 u = f  &\quad \text{ in } \Omega, \\[5pt]
\phantom{\det D^2} u=g & \quad \text{ on } \partial\Omega,
\end{array}
\right.
\end{align}
where $\Omega\subset\mathbb R^n$, $n\geq 1$, is an open bounded convex set,  and $f=f(x)$ and $g=g(x)$ are given functions defined on $\Omega$ and $\partial \Omega$ respectively. We require $f$ to be non-negative and locally integrable, and $g$ is assumed to be continuous. 
For the purpose of illustration we will focus on the problem in \eqref{prob:MA_D}, where $f=f(x)$, but generalizations to some of the above mentioned cases/applications where $f=f(x,\,u,\,\nabla u)$ may also be possible, and this will be the focus of future publications.

From the mathematical perspective, it is well-known that when $\Omega$ is strictly convex, then \eqref{prob:MA_D} has a unique convex solution in the sense of Alexandrov provided that the right-hand side is nonnegative and integrable \cite{RaTa}. In fact, one may even allow the right-hand side to be a finite nonnegative Borel measure. In the case of a merely convex domain, a unique convex solution exists if, in addition, the function $g$ can be extended to a convex function on $\bar{\Omega}$, see \cite{Ha}. However, it must be noted that also nonconvex solutions might exist. Even in the two-dimensional case, Rellich's Theorem \cite{Re} only states that there are at most two solutions. For an elementary illustration of nonuniqueness, consider the case in which the dimension of $\Omega$ is even and $g=0$. Then, for any convex solution $u$ we also have the concave solution $-u$. In general, the mathematical analysis of the Monge-Amp\`ere equation has been a source of intense investigations in the last decades and in Subsection \ref{MA} we give a brief account of  the Monge-Amp\`ere equation and provide references to the vast literature.

From the numerical perspective the problem in \eqref{prob:MA_D} is challenging for many reasons but in particular due to the presence of the fully nonlinear Monge-Amp\`ere operator and due to the fact that since the unique convex solution is typically relevant for applications, a numerical method has to ensure that it singles out this solution. Many different approaches/methods for the Dirichlet problem for the Monge-Amp\`ere equation have been proposed in the literature including the discretization of the determinant, wide stencil finite difference techniques, mixed finite-element methods, reformulation of the Monge-Amp\`ere problem as a Hamilton-Jacobi-Bellman equation, and many other methods. We
give a brief review of the numerical literature devoted to the Monge-Amp\`ere equation in Subsection \ref{numam}.

As the review in Subsection \ref{numam} underscores, there seems to be no literature devoted to  solving the Monge-Amp\`ere equation using neural networks. In this paper we start filling this gap by developing two methods for solving the Dirichlet problem in \eqref{prob:MA_D} using neural networks. The methods are illustrated in a number of examples and these examples show that neural networks are also promising tools for solving fully nonlinear equations like the Monge-Amp\`ere equation in complex geometries and in higher dimensions. In the context of neural networks, one way to attempt to single out the unique convex solution to the Monge-Amp\`ere equation  is to use a loss function which penalizes lack of convexity. While such a  method is easy to implement in the two-dimensional case, it is not necessarily extendible to higher dimensions. Another method, which is the main focus in this paper, is to use input convex neural networks. These networks are convex in the input variable by construction, so at least in principle they should provide a good ansatz for the solution to \eqref{prob:MA_D} in any dimension.

The rest of the paper is organized as follows. In Section \ref{prelim}, which mainly is of preliminary nature, we briefly discuss the theory for the Monge-Amp\`ere equation, we review the numerical literature devoted to the Monge-Amp\`ere equation, and recall the definition of fully connected feed-forward neural networks. The interested reader can find some additional background for the Monge-Amp\`ere equation in Appendix \ref{app:MA}. In Appendix \ref{app:approximation} we provide an overview of approximation results for neural networks, underscoring the importance of deeper neural networks as a vehicle to overcome the curse of dimensionality encountered by all other numerical schemes in higher dimensions.
In Section \ref{sec:Dirichlet} we present our methods in detail and make some comparison with previous work. In Section \ref{sec:NumEx} we present numerical examples of the solution of \eqref{prob:MA_D}. We first compare the two methods in the case of a smooth radial source function. In the other examples we focus on the input convex networks \cite{AmXuKo}, and consider functions with increasingly singular behavior on the boundary as well an example with a discontinuous source function. We also investigate how the method performs on a nontrivial domain and study its stability with respect to noise in the source function. Finally, we demonstrate that the method works in higher dimensions. In Section \ref{sec:numconvg} we study the numerical convergence of the method and in Section \ref{sec:error_est} we outline a method for estimating the error for sufficiently regular right-hand sides and boundary data.  The paper ends with a brief section devoted to a summary and conclusions. 

\vskip0.4cm \noindent 

\noindent
{\bf Acknowledgments.} This work was partially supported by the Wallenberg AI, Autonomous Systems and Software Program (WASP) funded by the Knut and Alice Wallenberg Foundation. Kaj Nystr\"om was supported  by the Swedish Research Council, dnr: 2022-03106.

\section{Preliminaries: the Monge-Amp\`ere equation, neural networks}\label{prelim}

In this section we give the reader some facts concerning the  theory for the Monge-Amp\`ere equation, we review the numerical literature devoted to the Monge-Amp\`ere equation, and we briefly discuss neural networks and the fundamental approximation problem.

\subsection{The Monge-Amp\`ere equation}\label{MA}  The mathematical analysis of the Monge-Amp\`ere equation has been a source of intense investigations in the last decades and we refer to the books \cite{CC,F,Gut} for detailed expositions. For the convenience of the reader, we here give a brief account concerning the well-posedness of the Dirichlet problem for the Monge-Amp\`ere equation mainly following Figalli \cite{F}.

To start we first note that if $v$ is a $C^2$ convex function on $\mathbb{R}^n$, and if $\Omega \subset \mathbb{R}^n$ is a domain, then the area formula gives for every Borel set $E\subset \Omega$,
\begin{align*}
\int_{E} \det D^2v \,\d x = |\nabla v(E)|,
\end{align*}
where $|\cdot |$ denotes the Lebesgue measure on $\R^n$. For an arbitrary, but not necessarily smooth, convex function $v$ on $\Omega$, and every Borel set $E \subset \Omega$, we define
\begin{align*}
Mv(E) := |\partial v(E)|,
\end{align*}
where $\partial v(E)$ is the set of slopes of supporting hyperplanes to the graph of $v$ (the sub-gradients of $v$) over points in $E$. It can be proved, see \cite[Theorem 2.3]{F}, that
\begin{align*}\label{MAMeasure}
\mbox{$Mv$ is a Borel measure on $\Omega$}.
\end{align*}
\noindent
$Mv$ is often referred to as the Monge-Amp\`{e}re measure of $v$ and if $v \in C^2$, then $Mv = \det D^2v\,\d x$, which shows that the concept of Alexandrov solution generalizes the classical notion of solution.  

\begin{defn}
Let $\mu$ be a Borel measure on a domain $\Omega \subset \mathbb{R}^n$. A convex function $u$ on $\Omega$ is said to be an Alexandrov solution of $\det D^2u = \mu$ if $Mu = \mu$.
\end{defn}

Alexandrov solutions are important and useful as they satisfy a maximum principle and as they have good compactness properties. Note that if $u$ and $v$ are convex on a bounded domain $\Omega$, with $u = v$ on $\partial \Omega$ and $u \leq v$ in $\Omega$, then
$$\partial v(\Omega) \subset \partial u(\Omega).$$
This is a simple consequence of convexity. Using this observation one concludes the following comparison principle \cite[Theorem 2.10]{F}.
\begin{thm}\label{ComparisonPrinciple}
Assume $u$ and $v$ are convex on a bounded domain $\Omega$, with $u \leq v$ on $\partial \Omega$. If $Mu \geq Mv$ in $\Omega$, then $u \leq v$ in $\Omega$.
\end{thm}
The comparison principle is useful for deducing the uniqueness of solutions to the Dirichlet problem, and we have made use of the result in Example \ref{surface_of_sphere} below. The following result has also been extensively used to motivate the existence and uniqueness of solutions in our numerical examples.

\begin{thm}\label{thmdp}
Let $\Omega \subset \mathbb{R}^n$ be a bounded strictly convex domain, let $\mu$ be a bounded Borel measure in $\Omega$, and let $g \in C(\partial \Omega)$. Then there exists a unique Alexandrov solution in $C(\overline{\Omega})$ to
the Dirichlet problem
\begin{align*}
\left\{
\begin{array}{ll}
\det D^2 u = \mu  &\quad \text{ in } \Omega, \\[5pt]
\phantom{\det D^2} u=g & \quad \text{ on } \partial\Omega,
\end{array}
\right.
\end{align*}
If $\Omega$ is merely convex, then the same conclusion holds provided that $g$ can be extended to a convex funciton on all of $\bar \Omega$.
\end{thm}
The proof for strictly convex $\Omega$ was originally published in \cite{RaTa}. See also the book \cite{F} for a rather streamlined approach. The extension to all convex $\Omega$ was proved in \cite{Ha}. A sketch of the proof in the for solutions vanishing on the boundary is provided in Appendix \ref{app:MA}, and we also briefly comment on how to extend the result to general $g$ in the strictly convex case.

\subsection{Numerical methods for the Monge-Amp\`ere equation}\label{numam} To give an account of the literature we note that spatial discretization of the Monge-Amp\`ere equation has mostly focused on the discretization of the determinant, e.g. see \cite{froese-oberman:2011-1}. In \cite{oberman:2008-1} a wide stencil finite difference technique is developed based
on the Barles and Souganidis framework \cite{BarS} to ensure discrete monotonicity guaranteeing convergence
of the numerical solution to the unique convex viscosity solution. Mixed finite-element methods have
been used by several authors including \cite{DG,feng-neilan:2009-1,LaP,Ne}. Finite differences have been used extensively in solving the Monge-Amp\`ere equation, e.g. see \cite{Delz}. In \cite{Browne,Budd} simple finite differences with filtering of the right hand side and smoothing of the
Hessian is used in order to ensure that a convex solution is found. Weller et al. \cite{Well} used finite volumes to
solve the Monge-Amp\`ere equation on a plane and on the sphere and explored a number of techniques, see also \cite{Fro} for some work on a wide stencil finite differences approach. In \cite{FJ}  a
reformulation of the Monge-Amp\`ere problem as a Hamilton-Jacobi-Bellman equation is given and removes the
constraint of convexity on the solution as a by-product of this reformulation. In \cite{BS} the authors solve the two-dimensional Dirichlet problem for the
Monge-Amp\`ere equation by a strong meshless collocation
technique that uses a polynomial trial space and
collocation in the domain and
on the boundary. In \cite{CaGloSo}, the Dirichlet problem for the Monge-Amp\`ere equation is solved in two dimensions using a least-squares formulation combined with a relaxation method. See also \cite{CaGloGo}, where the three-dimensional case is considered. A least squares method for optimal transport using the Monge-Amp\`ere equation was considered in \cite{PriBeIjTu}. Certain problems in optical design can be formulated using the Monge-Amp\`ere equation, and in these cases constructive methods using supporting surfaces can be used, see for example \cite{CaKoOl} for an analysis of the method in the far-field case. For a few other notable works we refer to \cite{BeFroeOb0,brenner-et-al:2011-1,froese-oberman:2013-1, li-liu:2017-1, liu-et-al:2017-1, liu-he:2013-1}.

While the literature outlined underscores the interest in solving the Monge-Amp\`ere equation, and attempts to ensure convergence to convex solutions, we note that none of these papers consider neural networks.

\subsection{Fully connected feed-forward neural networks} In this paper we consider two types of neural networks: standard feed-forward neural networks and input convex neural networks.  In this section we describe the standard fully connected feed-forward artificial neural networks, for which several approximation results are available, see for example Appendix \ref{app:approximation} below for a summary. The networks appearing in our first method, described in Subsection \ref{subsec:mod_loss}, are also of this type. The architecture of the input convex networks used in our second method is somewhat more involved, but it can also be seen as a special case of a feed-forward network. We refer to Subsection \ref{subsec:input_conv} for the details.

A standard feed-forward network is a function $f:\R^n \to \R^m$ of the form
\begin{equation*}
\label{feedforward---}
\begin{aligned}
z^{1} &= W^1x + b^1.\\
z^{2} &= W^2\sigma_1(z^1) + b^2 \\
& \mathrel{\makebox[\widthof{=}]{\vdots}} \\
z^{L-1} &= W^{L-1}\sigma_{L-2}(z^{L-2}) + b^{L-1} \\
z^L &= W^L\sigma_{L-1}(z^{L-1}) + b^L \\
y^L &= \sigma_L(z^L) \\
f(x) &= y^L\\
\end{aligned}
\end{equation*}
Here $W^j$ are matrices, $b^j$ are vectors and $\sigma_j$ are (typically non-affine) functions between euclidean spaces. The vectors and matrices must be chosen so that every $z^j$ has the dimension matching the domain of $\sigma_j$. 

A network with the architecture described above is said to have $L+ 1$ layers, where layer 0 is the input layer and layer $L$ is the output layer. The layers $0 < l < L$ are referred to as the hidden layers. The so-called activation functions $\sigma_j$ in the hidden layers are often defined by applying some non-affine function such as the sigmoid function, the rectified linear unit, or the hyperbolic tangent to each component of $z^j$. Not all activation functions are of this type, for example the softmax function frequently used in classification problems. In our case the activation function for the output layer will be the identity. In our context, we choose the dimensions of $W^j$ and $b^j$ so that the output is one-dimensional, i.~e. $f$ defines a mapping $\mathbb{R}^n \to \mathbb{R}$.


For the calibration of feed-forward artificial neural network and more general deep neural networks using back propagation we refer to \cite{difficultdeep,firstdeep,backprop,dropout2}. For more references and for a modern and modestly technical overview of several aspects of deep learning we refer to the popular book \cite{Goodfellow-et-al-2016}.


\section{Two methods to solve the  Monge-Amp\`ere equation using neural networks}\label{sec:Dirichlet}
In this section we present the two methods considered for solving problem \eqref{prob:MA_D} numerically. In both methods, the goal is to find an approximation for the unique convex solution by ensuring that the ansatz satisfies the equation pointwise and attains the correct boundary values on sufficiently densely spaced points. This in turn, is achieved by minimizing a loss function with respect to the parameters appearing in the neural networks. The main difference between the methods is how convexity of the solution is achieved. The first approach is inspired  by  \cite{BeNy} but based on a modified loss function
constructed by adding a term to the loss function which penalizes nonconvexity. The second approach uses an ansatz based on so-called input convex neural networks as introduced in \cite{AmXuKo}.

\subsection{Modified loss function}\label{subsec:mod_loss}
In \cite{BeNy} approximate solutions to Dirichlet problems of the form
\begin{align*}
\left\{
\begin{array}{ll}
L u = f  &\quad \text{ in } \Omega, \\[5pt]
\phantom{L}u=g & \quad \text{ on } \partial\Omega,
\end{array}
\right.
\end{align*}
where $L$ is a linear operator and $f$ and $g$ are given functions, were constructed by considering an ansatz of the form
\begin{align}\label{Ansatz:BeNy}
u(w,x) = G(x) + D(x)N(w,x),
\end{align}
where $G$ is a fixed smooth function satisfying the Dirichlet boundary condition, $D$ is a fixed smooth function which is positive inside the domain but vanishes on its boundary, and $N$ is a deep neural network. As explained in \cite{BeNy}, a function $G$ can be obtained by training a neural network to approximate $g$. The idea for constructing a suitable function $D$ is roughly speaking to approximate the distance function $x\mapsto d(x,\mathbb R^n\setminus \Omega)$ in $\Omega$ by a neural network. See \cite{BeNy} for details.

 A function $u$ of the form \eqref{Ansatz:BeNy} automatically satisfies the boundary condition, regardless of the form of $N$, so it only remains to choose the weights and biases $w$ so that the equation is satisfied. In \cite{BeNy} this is done by minimizing a discretized form of the expression
\begin{align*}
\int_\Omega |L u(x,w)- f(x)|^2 \d x.
\end{align*}
Thus, the loss function corresponding to the equation takes the form
\begin{align*}
E := \frac{1}{M}\sum^M_{j=1} |Lu(x_j,w)-f(x_j)|^2,
\end{align*}
where $\{x_j\}_{j=1}^M$ is a set of some sufficiently densely chosen collocation points in the domain $\Omega$.

To try to directly apply this approach to the Dirichlet problem for the Monge-Amp\`ere equation in \eqref{prob:MA_D} is not a good idea as there are no guarantees that the network will converge to the unique convex solution rather than some nonconvex solution. However, one can attempt to remedy this problem by adding a term to the loss function which penalizes lack of convexity. In the following we will explain how such a term can be constructed in the two-dimensional case. Recall that $u$ is convex if and only if its Hessian is positive semi-definite, which is true if and only if the eigenvalues of the Hessian are nonnegative. In the two-dimensional case, the eigenvalues of a symmetric matrix $H = [h_{ij}]$ can be calculated explicitly and are given by
\begin{align}\label{lamb}
\lambda = \frac{h_{11}+h_{22}\pm \sqrt{(h_{11}+h_{22})^2-\det H}}{2}.
\end{align}
As we are including a term in the loss function coming from the equation, namely \footnote{The derivatives appearing on the right-hand side of \eqref{eq_loss} are calculated using automatic differentiation in TensorFlow, see Section \ref{sec:NumEx}.}
\begin{align}\label{eq_loss}
E_e = \sum^M_{j=1} |\det D^2 u(x_j,w)-f(x_j)|^2, 
\end{align}
we are by construction, as the function $f$ is assumed to be nonnegative, training the network so that the Hessian of $u$ has nonnegative determinant. Thus, the expression for $\lambda$ in \eqref{lamb} reveals that the eigenvalues are nonnegative if and only if $h_{11}+h_{22} \geq 0$. This holds for example if both $h_{11}$ and $h_{22}$ are nonnegative. Conversely, the diagonal entries of a positive semi-definite matrix must always be nonnegative. Thus, we see that the matrix is positive semi-definite if and only if both $h_{11}$ and $h_{22}$ are nonnegative. This prompts us to add a term proportional to
\begin{align}\label{conv_loss}
E_c := \frac{1}{M}\sum^M_{j=1}( \min\{0, \partial_x^2 u(x_j,w)\}^2 + \min\{0, \partial_y^2 u(x_j,w)\}^2 )
\end{align}
to the loss function. Hence, in this approach we take the loss function $E$ as a suitably weighted sum of the terms \eqref{eq_loss} and \eqref{conv_loss},
\begin{align}\label{tot_loss:method1}
E := E_e + CE_c,
\end{align}
where $C$ is a suitably large constant.

Inspired by Brix, Hafizogullari and Platen \cite{BriHaPla,BriHaPla2} one could  in addition replace the determinant of the Hessian in \eqref{eq_loss} by the quantity
\begin{align}\label{Modd}
\det H^+_u(x_j,w) := \max\{0, \partial_x^2 u(x_j,w)\} \max\{0, \partial_y^2 u(x_j,w)\} - (\partial_x\partial_y u(x_j,w))^2.
\end{align}
In the case of a strictly positive $f$, this guarantees that also the loss coming from the equation is large whenever one of the diagonal elements is nonpositive.  However, in our case we found no improvement in accuracy using this approach. One reason might be that training seems to be fastest with a relatively high value for the constant $C$ in front of the convexity term $E_c$, thus rendering the modification of the determinant less significant. In our numerical experiments, using the original determinant turned out to require fewer iterations for convergence  compared to using the quantity in \eqref{Modd}.

In any case, the presented modification of the loss function, to ensure convexity, essentially only works in the two-dimensional case, since in higher dimensions it becomes harder to check the sign of the eigenvalues. Therefore in higher dimensions different approaches are called for.

\subsection{Input convex networks}\label{subsec:input_conv}
 A way to ensure the convexity of $u$ which is independent of dimension, is to use so-called input convex neural networks as introduced in \cite{AmXuKo}. The architecture of this type of network guarantees convexity. The general form of an input convex neural network $N(\theta,x)$ with a total of $L+1$ layers, input variables $x$ and parameters $\theta$ is given by
\begin{align}\label{architecture:input_conv}
z_1 &= g_0(L_0 x + b_0),
\\
\notag z_{j+1} &= g_j(W_j z_j + L_j x + b_j), \textrm{ for } j \in \{1, \dots, L-1 \},
\\
\notag N(\theta,x) &= z_L.
\end{align}
Here, $\{g_j\}^L_{j=0}$ are given increasing convex functions which act component-wise on vector valued input, $W_i$ are matrices with nonnegative entries, $L_j$ are matrices and $b_j$ are vectors. Hence, $\theta$ refers to all the parameters $\{\{L_j\}^{L-1}_{j=0}, \{W_i\}^{L-1}_{i=1}, \{b_j\}^{L-1}_{j=0}\}$. The parameters for the last layer are chosen so that the output $z_L$ is one-dimensional. The convexity of the network $N(\theta,x)$  follows from the following two facts:
\begin{enumerate}
\item A linear combination with positive coefficients of convex functions is convex.

\item The composition $h \circ g$ of a convex function $g:\R^n \to \R$ and an increasing convex function $h:\R\to \R$ is convex.
\end{enumerate}

It is in principle possible to describe input convex networks using the framework of standard feed-forward networks, if one uses activation functions of the form $\textnormal{id}_{\R^n}\times G$ where $n$ is the input dimension and $G = g\times \dots \times g$ where $g:\R\to \R$ is convex. Taking also matrices $W_j$ and vectors $b_j$ so that the $n$ first components are preserved in the affine transformations we could ensure that in every layer (except the final layer) the $n$ first components  of the output are $x$ and that the remaining components behave as in \eqref{architecture:input_conv}. Although interesting, it is unclear whether this point of view provides any additional insight.

When using input convex neural networks to solve \eqref{prob:MA_D}, our ansatz $u$ will consist of a single input convex neural network, rather than the more involved ansatz \eqref{Ansatz:BeNy} used previously. The reason is that even if we chose the  networks $G$, $D$ and $N$ in \eqref{Ansatz:BeNy} to be convex, the product $DN$ might fail to be convex, which would defeat the goal. Therefore, we can no longer pre-train the ansatz to satisfy the Dirichlet boundary condition. Instead, the boundary condition is taken into account by adding a term of the form
\begin{align*}
E_b:= \frac{1}{K}\sum^K_{k=1} |u(y_k,w)-g(y_k)|^2,
\end{align*}
to the loss function, where the points $y_k$ are some relatively densely distributed points on $\partial \Omega$. Thus we are able to omit the convexity term in the loss function at the expense of including a loss term corresponding to the Dirichlet boundary condition. The loss function used in this case is therefore
\begin{align}\label{tot_loss:method2}
E:= E_e + C E_b,
\end{align}
for a suitable constant $C$.

\section{Numerical examples}\label{sec:NumEx}
In this section we apply the two methods introduced in the previous section to concrete examples. In our numerical calculations, all derivatives, such as the second order derivatives occuring in the Hessian, and the gradient of the loss functions with respect to the weights, have been calculated using automatic differentiation in TensorFlow \cite{TensorFlow}. We train the networks, i.e. minimize the loss functions, using the BFGS algorithm from the SciPy library \cite{SciPy}. This algorigthm only makes use of function evaluations and first order derivatives, and terminates in the SciPy implementation when the euclidean norm of the gradient becomes sufficiently small (we used the default value 1.0e-5). A custom implementation of the Adam algorithm making use of TensorFlow has also been used.

Initial values for weights and biases are chosen randomly according to the normal distribution. In the case of input convex networks, certain parameters $w$ need to remain nonnegative. We have solved this issue by writing such parameters in the form $w=v^2$ and we train w.r.t. $v$ rather than $w$.

In order to generate the randomly selected collocation points we have used the method numpy.random.uniform from the NumPy library. For example, for the unit square we pick pairs of uniformly distributed points on $[0,1)$. In order to generate points on the boundary which lie on some line segment (for example the sides of the unit square) we again use the same method and map the obtained points on $[0,1)$ onto the line segment by an affine transformation. For the more complicated domain in Example \ref{assym_domain}, one part of the boundary lies on the boundary of a circle, and here we use a trigonometric parametrization to map randomly selected points on the interval to the boundary. In order to obtain points inside the domain of Example \ref{assym_domain}, we select random points on the unit square, rejecting any points that happen to lie outside the domain. 

In the first example we solve problem \eqref{prob:MA_D} for a given source function $f$ and boundary values $g$ using the method described in Section \ref{subsec:mod_loss}, i.e. by penalizing nonconvexity. In the second example we solve the very same problem using input convex networks, allowing us to compare the two methods. In the subsequent two examples we investigate how the method with input convex networks performs when the source function has a singularity on the boundary. We also consider a case in which the right-hand side is discontinuous. In further examples we turn our attention to a nontrivial domain and we also investigate the effect of noise in the source term. Finally, we use input convex networks to solve higher-dimensional equivalents of the problem considered in the first two examples.

Some of the two-dimensional examples on the unit square in this section were previously considered in \cite{BeFroeOb0}, where the authors used both a finite difference method and a method based on iteratively solving a Poisson equation. We have not attempted to reproduce the method in \cite{BeFroeOb0} and we refer the interested reader to \cite{BeFroeOb0} for comparison.

When testing the methods it is important that we consider problems for which a solution exists and for which the solution is unique. We also need to know the exact expression for the solution so that we can compare it with the results of the numerical methods. In each example, except Example \ref{discont_source}, an exact solution to the Monge-Amp\`ere equation with a certain right-hand side has been constructed by substituting a convex $C^2$ function $u$ into the left-hand side of the equation and taking the result as the right-hand side $f$. In Example \ref{discont_source} the construction of the solution is a bit more involved, as there are points for which the second derivatives are not defined, and in this case we need to calculate the subdifferential in order to conclude that the function is indeed an Alexandrov solution with a specific right-hand side. Taking $g$ as the restriction of $u$ to the boundary of the domain $\Omega$ we have thus constructed a solution to the Dirichlet problem for the Monge-Amp\`ere equation with right-hand side $f$ and boundary values $g$.
Uniqueness of the solution in the class of convex functions follows from Theorem \ref{ComparisonPrinciple}. Since we construct explicit solutions, we can in principle avoid using Theorem \ref{thmdp}, but it is interesting to note that in all examples except Example \ref{surface_of_sphere}, the existence and uniqueness of a solution with the given right-hand side $f$ and boundary data $g$ could also be concluded using Theorem \ref{thmdp}. 

\subsection{Radial smooth source. Modified loss approach}\label{subsec:radial0}
We consider the smooth radial function
\begin{align}\label{radial:2dim}
u(x,y) = \exp\Big(\frac{x^2+y^2}{2}\Big),
\end{align}
on the unit square $\Omega = (0,1)^2$. A direct calculation shows that
\begin{align}\label{f-rhs}
f(x,y) := \det D^2 u(x,y) = (1+x^2+y^2)\exp(x^2+y^2).
\end{align}
Let $g$ denote the restriction of $u$ to $\partial \Omega$. By construction, $u$ is a convex solution to the Dirichlet problem  in \eqref{prob:MA_D} with right-hand side $f$ defined by \eqref{f-rhs} and boundary values $g$. Uniqueness of the solution in the class of convex functions follows from the comparison principle of Theorem \ref{ComparisonPrinciple}. Existence of a solution with boundary values given by \eqref{radial:2dim} could also be concluded from the second part of Theorem \ref{thmdp}  since evidently the expression in \eqref{radial:2dim} is a convex function on all of $\bar \Omega$. 

In our numerical illustrations we  use 3000 randomly distributed collocation points inside the domain, and 400 uniformly distributed points on the boundary.

To construct $G$ we approximate $g$ with a neural network with one hidden layer and 20 hidden nodes using the BFGS algorithm, and the mean squared error as loss. We use the hyperbolic tangent as activation function on the hidden layer, but no activation function on the output layer.

To construct the smooth approximate distance function $D$ we follow the approach of \cite{BeNy} with one simplification: since we are considering the unit square, the exact distance function is known, and we use it to calculate the exact distance to the boundary at 300 randomly chosen points in the interior of $\Omega$. We then fit a neural network $D$ to this data, along with the 400 data points corresponding to the boundary (where the distance should naturally be zero). We use one hidden layer, 40 nodes and the hyperbolic tangent as activation function on the hidden layer. No activation function was used on the output layer. Again, BFGS was used to train the network.

\begin{figure}
\includegraphics[scale=0.6]{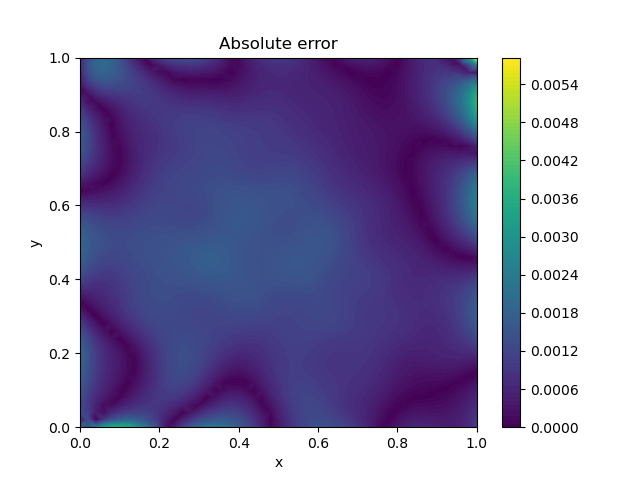}
\caption{Absolute error for Example \ref{subsec:radial0}}\label{Ex0_pic}
\end{figure}

The deep network $N(w,x)$ is constructed with five hidden layers with sigmoid as activation function on the hidden layers. Each hidden layer is taken to have 10 nodes, leading to a total of 481 parameters. We use no activation function on the output layer, as we want the possible range of the network to be all of $\R$.  We use the BFGS algortithm to train the network. In the loss function \eqref{tot_loss:method1}, we take the factor $C=10^4$ in front of the term which penalizes non-convexity. At some point the algorithm terminated due to precision loss. This happens when the line search in the BFGS method fails and could be a consequence of the local behavior of the function. In these cases we found that continuing to train the network using the Adam algorithm for a few hundred epochs with a learning rate of the order $10^{-7}$ took us out of the problematic region and allowed us to continue using the BFGS algorithm. The end result can be seen in Figure \ref{Ex0_pic}, which displays the absolute error of the trained network and the exact solution $u$. Note that by construction $u$ is of the order of 1 on the unit square. The precision obtained is comparable to that found in the examples in \cite{BeNy}.

\subsection{Radial smooth source. Input convex network.}\label{subsec_radial1}
In this example we solve the problem from the previous example using the method of Section \ref{subsec:input_conv}. We use an input convex network with five hidden layers and 10 nodes for each hidden layer. The total number parameters (weights and biases) in this case  is 563. We use softplus as the activation function on the hidden layers. This is a convex increasing function, as required by the method, and we note that it is also smooth, thus giving rise to a smooth network. We do not use any activation function on the output layer in order not to restrict the possible range of the network. For example, the range of softplus is $(0,\infty)$ and in general solutions may very well take negative values. For the model case studied here we could have used softplus as activation function also for the output layer since we know that the exact solution $u$ is non-negative. However, we want the method to be applicable also when such additional information about the unknown solution is not available.

\begin{figure}
\includegraphics[scale=0.6]{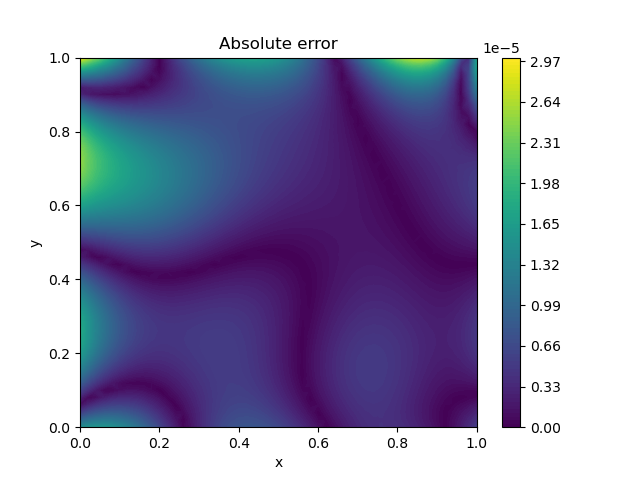}
\caption{Absolute error for Example \ref{subsec_radial1}}\label{Ex1_pic}
\end{figure}

As in the previous case, we took 3000 points inside the domain and 400 points on the boundary. In the term $E_e$ of the loss function \eqref{tot_loss:method2} we have also included the boundary points, and we take $C = 100$ as the factor in front of the boundary loss $E_b$. We trained the network using the BFGS algorithm, and obtain convergence after 2287 iterations. Figure \ref{Ex1_pic} shows the absolute error between the trained network and the exact solution \eqref{radial:2dim}. It is interesting to note, by comparing  Figure \ref{Ex0_pic} and  Figure \ref{Ex1_pic}, that the absolute error, and by construction then the relative error, that in Figure \ref{Ex0_pic} errors are of the order $10^{-3}$ while in Figure \ref{Ex1_pic} errors are of the order $10^{-5}$.

\subsection{Blow-up of the source function}\label{source_blowup1}
\begin{figure}[h]
\includegraphics[scale=0.6]{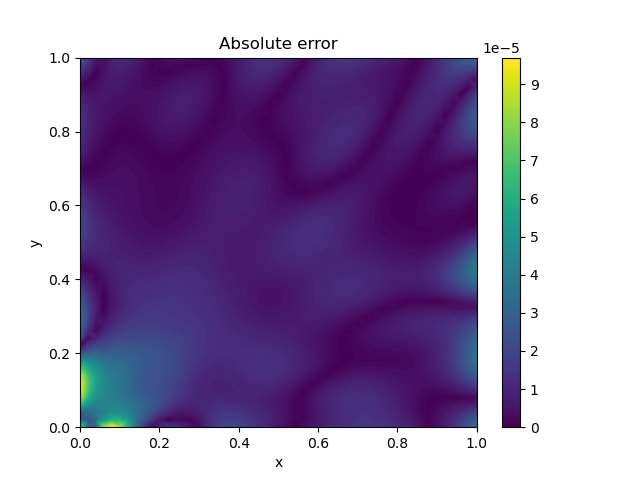}
\caption{Absolute error for Example \ref{source_blowup1}}\label{Ex2_pic}
\end{figure}

We consider the function
\begin{align*}
u(x,y) = \tfrac{2\sqrt{2}}{3}(x^2+y^2)^\frac{3}{4},
\end{align*}
on the unit square. A direct calculation shows that $u$ solves the Monge-Amp\`ere equation with the source function
\begin{align*}
f(x,y) = \det D^2u(x,y)  = \frac{1}{(x^2+y^2)^\frac{1}{2}}.
\end{align*}
Note that $f$ has a singularity at the origin, which lies on the boundary of the domain. Uniqueness of the solution to the Dirichlet problem with boundary values $g= u|_{\partial \Omega}$ follows from Theorem \ref{ComparisonPrinciple}. Since the right-hand side $f$ is integrable, one could also have concluded the existence of a solution from Theorem \ref{thmdp}. We found that the method from Example \ref{subsec_radial1} works also in this case, but that it can be improved by adding more points to the boundary loss near the singularity. These points are also given a higher weight. Thus, the loss function we use takes the form
\begin{align}\label{tot_loss:method2_singular}
E = E_e + C E_b + K E_B,
\end{align}
where $E_b$ is the original boundary loss corresponding to randomly distributed points on the   boundary, and
\begin{align*}
E_B:= \frac{1}{P}\sum^P_{k=1} |u(z_k,w)-g(z_k)|^2,
\end{align*}
where $\{z_k\}$ are some points lying on the boundary near the singularity. In our case we took 20 equidistantly distributed points for each of the line segments $[0, 0.04] \times \{0\}$ and $\{0\} \times [0, 0.04]$. In \eqref{tot_loss:method2_singular} we took $C=1000$ and $K = 4000$. The addition of the term $E_B$ to the loss function allows faster convergence: after 5000 iterations the absolute error was already smaller than what we ever could obtain without the extra loss term. In addition we obtain higher accuracy, as the maximum error was reduced by more than $90 \%$. The absolute error after 55000 iterations can be seen in Figure \ref{Ex2_pic}.

\subsection{Surface of a sphere}\label{surface_of_sphere}
We now consider the example
\begin{align*}
u(x,y) = -\sqrt{2-x^2-y^2}, \hspace{7mm} f(x,y):= \det D^2 u(x,y) = \frac{2}{(2 - x^2 - y^2)^2}.
\end{align*}
In this case the graph of $u$ describes a part of the surface of a sphere. The right-hand side $f$ has a singularity at the point $(1,1)$ which lies on the boundary of the domain. Note that the singularity is stronger than  in Example \ref{source_blowup1}, and $f$ is not integrable in this example. Thus, existence of a unique solution could not be concluded using Theorem \ref{thmdp}. However, since we have the explicit solution $u$ given above, and the comparison principle of Theorem \ref{ComparisonPrinciple} is still valid, we know that there is precisely one solution.

\begin{figure}[h]
\includegraphics[scale=0.6]{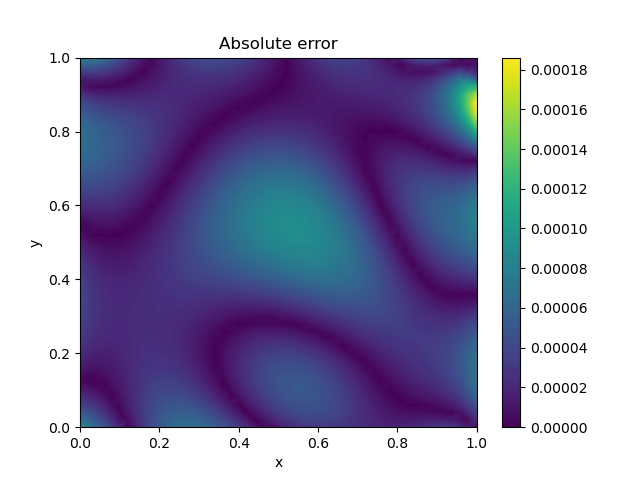}
\caption{Absolute error for Example \ref{surface_of_sphere}}\label{Ex3_pic}
\end{figure}

We apply the method from the previous example with the same values as before for $C$ and $K$ in \eqref{tot_loss:method2_singular}. In this case, we choose 20 extra points for each of the line segments $[0.95,1]\times\{1\}$ and $\{1\}\times [0.95,1]$. The BFGS algorithm terminated due to precision loss on a number of occasions, most likely due to the strongly singular source function. In these cases a few iterations with the Adam algorithm allowed us to escape the problematic region and to continue training with BFGS. Figure \ref{Ex3_pic} shows the end result.

\subsection{Discontinuous source function}\label{discont_source}
\begin{figure}
\includegraphics[scale=0.6]{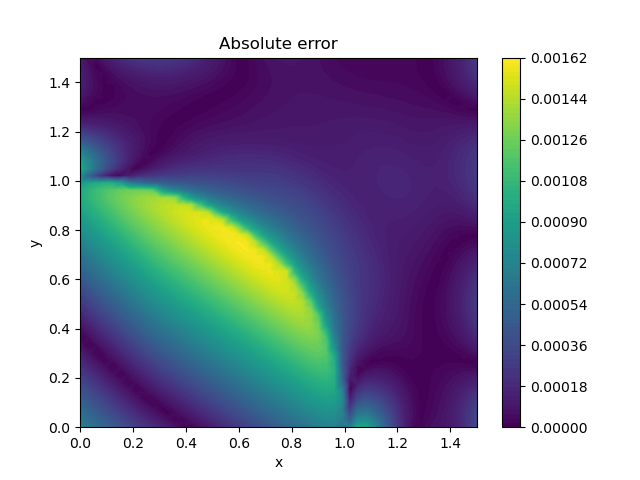}
\caption{Absolute error for Example \ref{discont_source} after 150000 iterations with 3000 points inside the domain and 400 points on the boundary.}\label{Ex10_pic1}
\end{figure}

We consider the square domain $\Omega = (0, 1.5)^2$ and the following convex function $u$ written in radial form as
\begin{align*}
 u(r) = 
 \begin{cases}
\, 0, & r \leq 1
\\
\, r \sqrt{r^2 -1} - \ln ( r + \sqrt{r^2 -1}), & r > 1.
\end{cases}
\end{align*}
An explicit calculation shows that
\begin{align*}
 \partial_r u(r) = 
 \begin{cases}
\, 0, & r \leq 1
\\
\, 2 \sqrt{r^2 -1}, & r > 1,
\end{cases}
\end{align*}
and we can conclude that $u$ is continuously differentiable. Moreover, since the determinant of the Hessian of a radial function takes the form $\frac1r (\partial_r u) (\partial_r^2 u)$, we see that $u$ satisfies the Monge-Amp\`ere equation with the right-hand side
\begin{align}\label{f_discont}
 f(r) = 
  \begin{cases}
\, 0, & r \leq 1
\\
\, 4, & r > 1,
\end{cases}
\end{align}
pointwise except at points for which $r=1$.  The form of $\partial_r u$ shows that these points do not make any contribution to the Monge-Amp\`ere measure, and we conclude that $u$ is an Alexandrov solution with the right-hand side \eqref{f_discont}. Moreover, uniqueness again follows either by Theorem \ref{thmdp} or directly from the comparison principle of Theorem \ref{ComparisonPrinciple}. We used the same network architecture as in Example \ref{subsec_radial1} and in our first calculations we chose 3000 points inside the domain and 400 points on the boundary. The end result after 15000 iterations can be seen in Figure \ref{Ex10_pic1}. 
The accuracy is significantly worse than previous examples, and the error is largest near the discontinuity points of the right-hand side, i.e. the points for which $r=1$.

\begin{figure}[hbt!]
\includegraphics[scale=0.6]{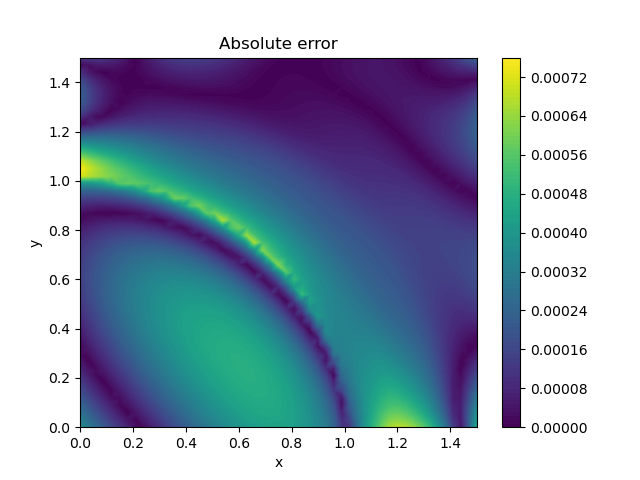}
\caption{Absolute error for Example \ref{discont_source} after 150000 iterations with 10000 points inside the domain and 1000 points on the boundary.}\label{Ex10_pic2}
\end{figure}

We first attempted to improve the accuracy by increasing the number of points in a small neighborhood of $r=1$. However we found that this approach prevents efficient training of the network. This is not very surprising as points close to the unit circle must be mapped to radically different values (4 or 0) depending on whether they are inside the circle or outside. This makes it hard to train the network. We found that increasing the number of points in the whole set $\Omega$ increases accuracy. Figure \ref{Ex10_pic2} shows the end result after 15000 iterations when we take 10000 points inside the domain and 1000 points on the boundary. As can be seen, the maximum error has been reduced to less than a half of the maximum error in Figure \ref{Ex10_pic1}. Still the error is fairly large compared to previous examples.

\pagebreak
\subsection{Asymmetric domain and the effect of noise}\label{assym_domain}
\begin{figure}
\includegraphics[scale=0.6]{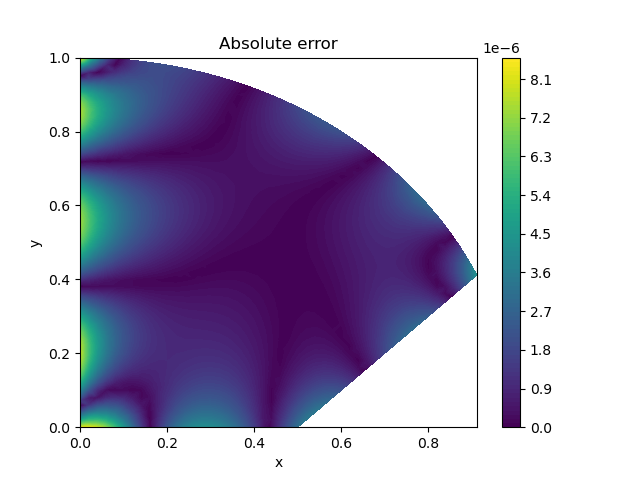}
\caption{Absolute error for Example \ref{assym_domain} without noise.}\label{Ex9_pic}
\end{figure}

In order to investigate how the method performs on less trivial domains than the unit square, in this example we solve the Dirichlet problem on the convex domain bounded by the conditions $x > 0$, $y > 0$, $x^2 + y^2< 1$ and $y > x - \tfrac12$. The domain can be viewed in Figure \ref{Ex9_pic}. As solution, we pick the smooth convex function
\begin{align*}
 u(x,y) = \tfrac{7}{10}\exp\Big(\tfrac12(x-\tfrac12)^2 + y^2\Big) + x^2 -\tfrac12.
\end{align*}
The choice is rather arbitrary, and our intention was to pick a function without any particular symmetries. We take 3000 points inside the domain and 400 points on the boundary. The number of points on each line or curve segment of the boundary is taken to be roughly proportional to its path length. We use the same network architecture as in Example \ref{subsec_radial1}. The result after 5000 iterations of BFGS can be seen in Figure \ref{Ex9_pic}.

We also investigated how noise in the source term affects the accuracy for this particular problem. In order to do this, we solved the problem where the values of the right-hand side have been modified by adding normally distributed noise, with standard deviations ranging from $10^{-3}$ to $1$. In all the cases we trained the network using BFGS until convergence or at most 10,000 iterations. In order to calculate the maximum and average error we compared the values of the network with the values of the actual solution at $10^6$ randomly chosen points inside the domain. The outcome of these calculations are presented in Table  \ref{table_noise_effect}. As can be seen, the method is rather stable with respect to noise. Somewhat surprisingly, the accuracy does not seem to change much in the interval $10^{-3}$ to $10^{-2}$. The obtained values can vary a bit on different runs due to the randomness in our methods, but the general trend is that noise decreases accuracy.

\begin{table}[h]
\begin{tabular}{|c|c|c|}
stdev & max error & average error \\
\hline
0 & 8.58e-06 & 1.17e-06 \\
$10^{-3}$ & 1.32e-05 & 2.57e-06 \\
$10^{-2}$ & 2.54e-05 & 3.64e-06 \\
$10^{-1}$ & 1.21e-04 & 2.67e-05 \\
1 & 2.48e-03 & 4.43e-04 \\
\end{tabular}\vspace{1mm}\caption{The effect of noise on the error in Example \ref{assym_domain}.}\label{table_noise_effect}
\end{table}

\subsection{Higher-dimensional examples}
Finally, we consider some higher-dimensional versions of Example \ref{subsec_radial1}. For example, in three dimensions we have
\begin{align*}
u(x,y,z) = \exp\Big(\frac{x^2+y^2+z^2}{2}\Big),
\end{align*}
defined on the unit cube. When constructing the loss function we take 4000 points inside the cube and 1200 points on the boundary. The BFGS algorithm was used in the optimization. Evidently we cannot plot the absolute error as a function in the three-dimensional cube, so instead we present the maximum error, average absolute error, and $L^2$-error in Table \ref{table_3dim}. The optimization terminated successfully after 20015 iterations.

\begin{table}[h]
\begin{tabular}{|c|c|c|c|}
iterations & max error & average error & L2 error \\
\hline
5000 & 4.029e-04 &  4.193e-05 & 6.079e-05 \\
10000 & 1.584e-04 & 9.967e-06 & 1.538e-05 \\
15000 & 8.801e-05 & 7.237e-06 & 1.115e-05 \\
20015 & 5.333e-05 & 5.250e-06 & 8.123e-06\\
\end{tabular}\vspace{1mm}\caption{The error for the three-dimensional radial function.}\label{table_3dim}
\end{table}

In the four-dimensional case we take 4000 points inside the cube and 2400 points on the boundary. The results are summarized in Table \ref{table_4dim}. The BFGS optimization terminated after 17506 due to precision loss.

\begin{table}[h]
\begin{tabular}{|c|c|c|c|}
iterations & max error & average error & L2 error \\
\hline
5000 & 4.275e-02 & 2.596e-03 & 3.893e-03 \\
10000 & 7.623e-03 & 4.462e-04 & 6.205e-04 \\
15000 & 2.895e-03 & 1.312e-04 & 1.871e-04 \\
17506 & 1.553e-03 & 7.387-05 & 1.090e-04 \\
\end{tabular}\vspace{1mm}\caption{The error for the four-dimensional radial function.}\label{table_4dim}
\end{table}

Real challenges start to occur in the five-dimensional case. We were unable to apply BFGS directly as the algorithm terminated immediately due to precision loss. With the Adam algorithm we were able to cause some decrease in the loss, but convergence was extremely slow. The only working approach was the following:
\begin{enumerate}
\item Train using Adam by only minimizing the boundary loss $E_b$ for up to 1000 epochs.
\item Train using Adam w.r.t. the full loss function for some thousands of epochs. This causes much quicker decrease in the loss than if step 1 was omitted.
\item Train with BFGS. At this point BFGS gives a rapid decrease in the loss, but terminates again due to precision loss.
\item Alternate between Adam and BFGS.
\end{enumerate}

When using Adam, we found that the single-batch approach was most efficient. The original loss in our case was of the order $10^7$. Using the method presented previously, we can decrease the loss to the order of $1$ rather quickly, but after this convergence becomes slow. When the loss function is this large, the approximate solution still differs significantly from the exact solution. Further development of the method is needed in order to obtain faster convergence and higher accuracy in this case.

\section{Numerical convergence}\label{sec:numconvg}
In this section we study the convergence of the method empirically, as the number of collocation points increases. We consider three different cases: increasing the number of interior points with a fixed number of boundary points, increasing the number of boundary points with a fixed number of interior points and finally increasing both the boundary points and interior points simultaneously. We examine three different examples: the boundary value problems from Examples \ref{subsec:radial0} and \ref{source_blowup1}, as well as the boundary value problem satisfied by the function in \eqref{radial:2dim} on the unit disk. In order to clearly separate the effect of boundary and interior points we have only included the interior points in the term $E_e$ in the loss function.

\subsection{Increasing interior points} 
\begin{figure}
\includegraphics[scale=0.6]{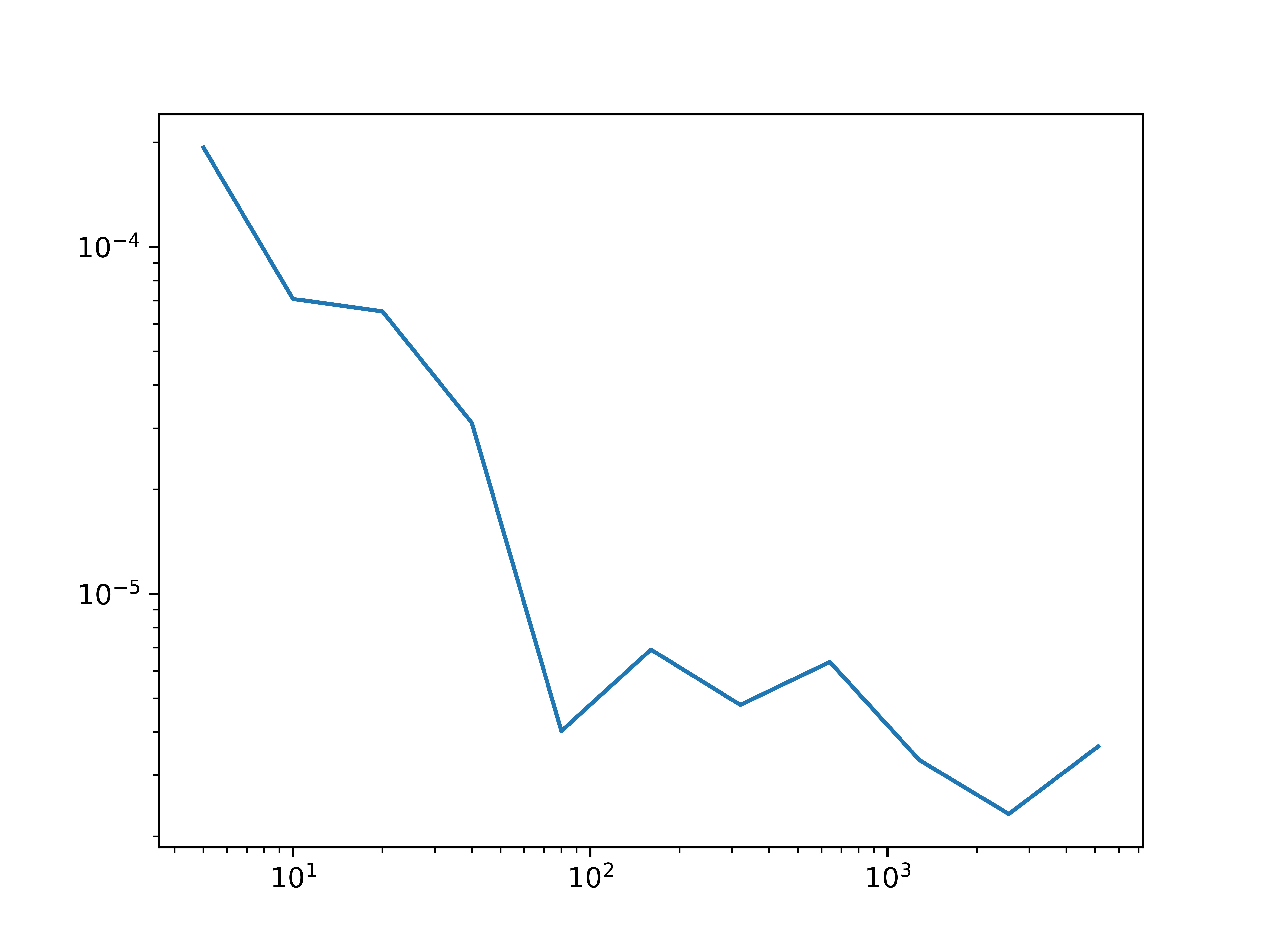}
\caption{The average absolute error as a function of the number of interior points for the boundary value problem of Example \ref{subsec:radial0}.}\label{IPI_rad_square}
\end{figure}

\begin{figure}
\includegraphics[scale=0.6]{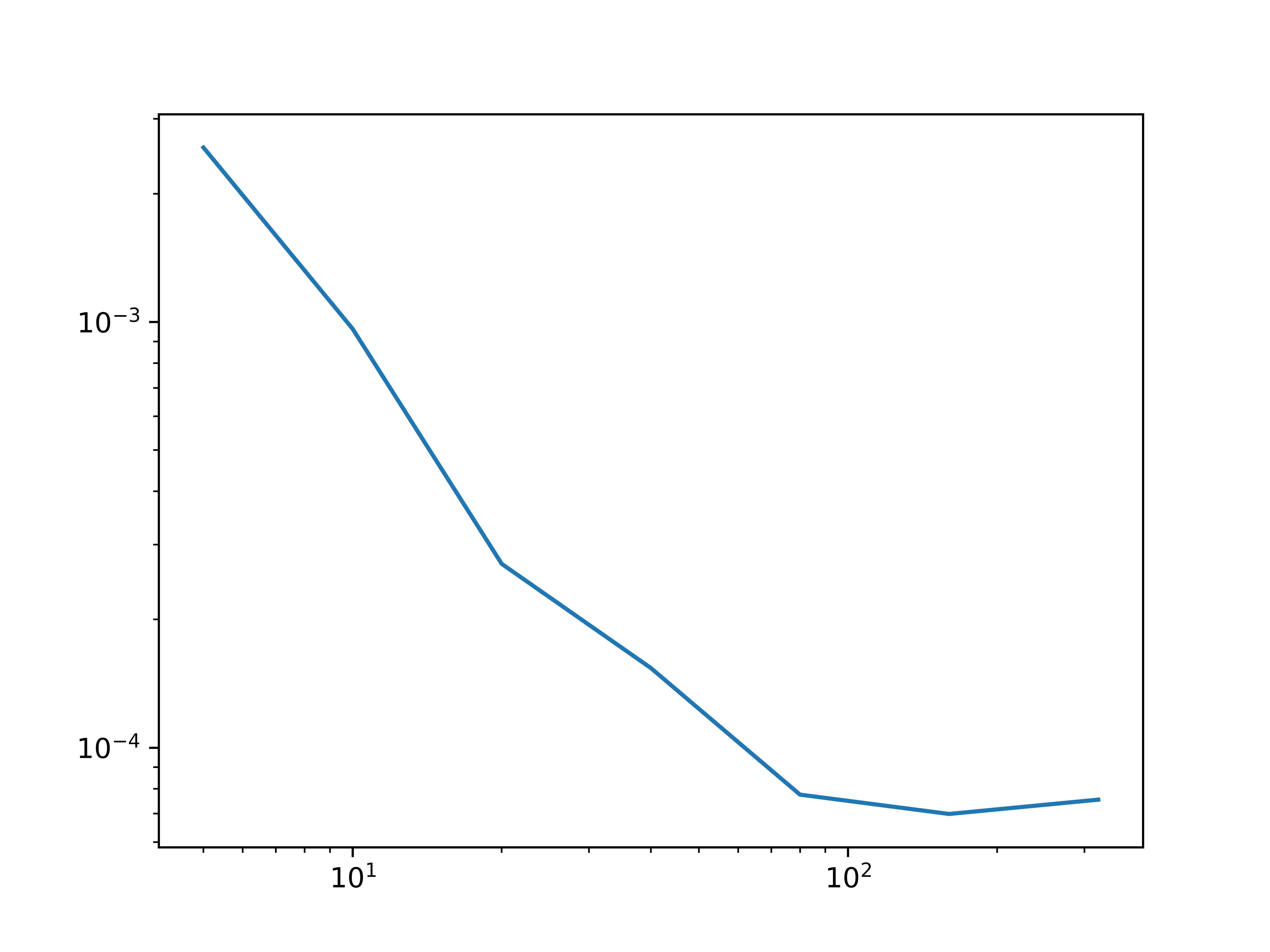}
\caption{The average absolute error as a function of the number of interior points for the boundary value problem of Example \ref{source_blowup1}.}\label{IPI_blowup}
\end{figure}

\begin{figure}
\includegraphics[scale=0.6]{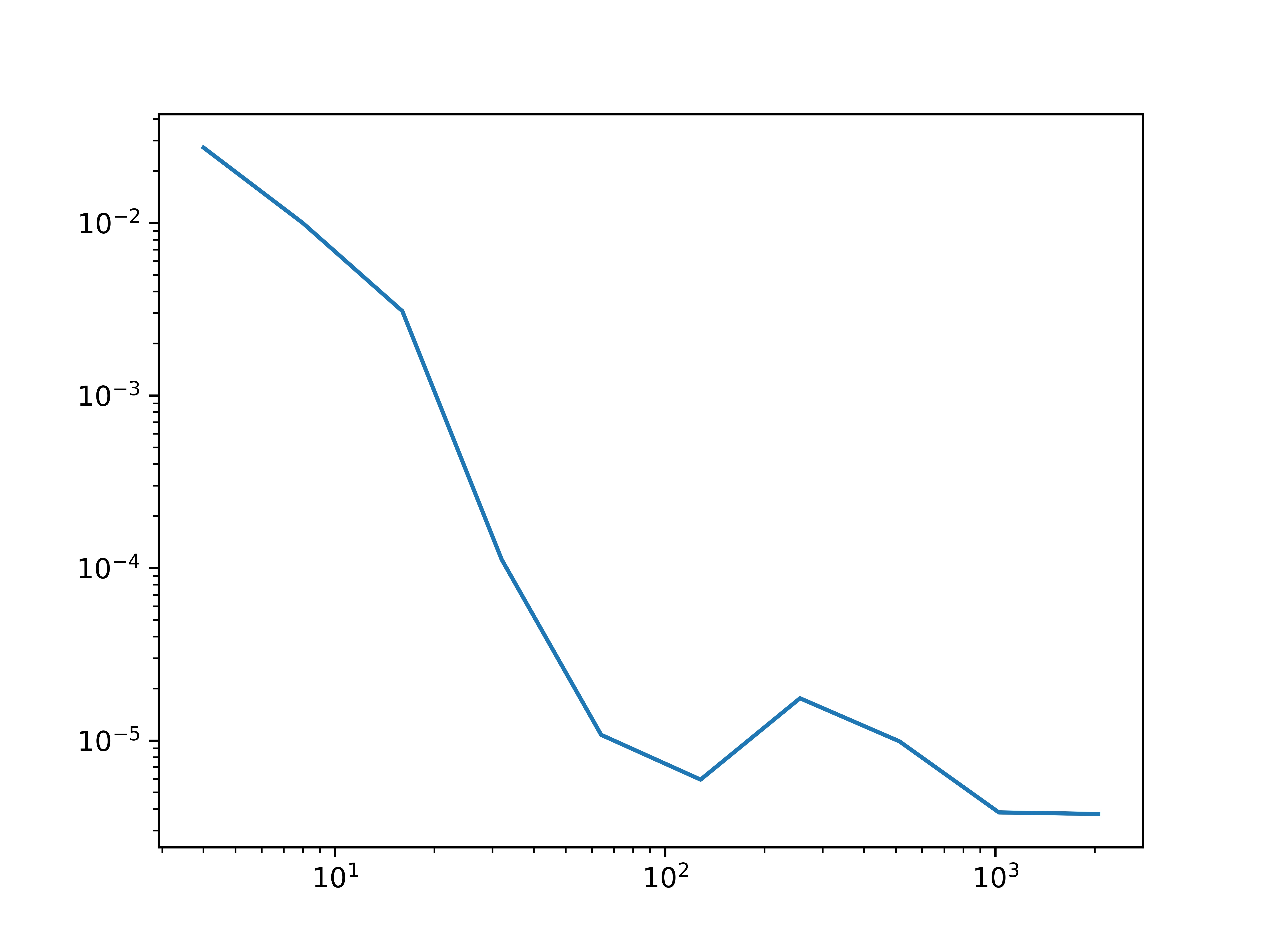}
\caption{The average absolute error as a function of the number of interior points for the boundary value problem corresponding to the function in \eqref{radial:2dim} on the unit disk.}\label{IPI_rad_disk}
\end{figure}

We fix the number of boundary points to be 400 in all examples but consider an increasing number of interior points. We train the network using BFGS until convergence or at most 5000 iterations. We start with only 4 or 5 interior points and double the amount in each computation until no major improvements can be seen. The average error as a function of the number of interior points for the three examples can be seen in Figures \ref{IPI_rad_square}, \ref{IPI_blowup} and \ref{IPI_rad_disk} respectively. Note that we have used a log-log scale. 

\subsection{Increasing boundary points}
\begin{figure}[hbt!]
\includegraphics[scale=0.6]{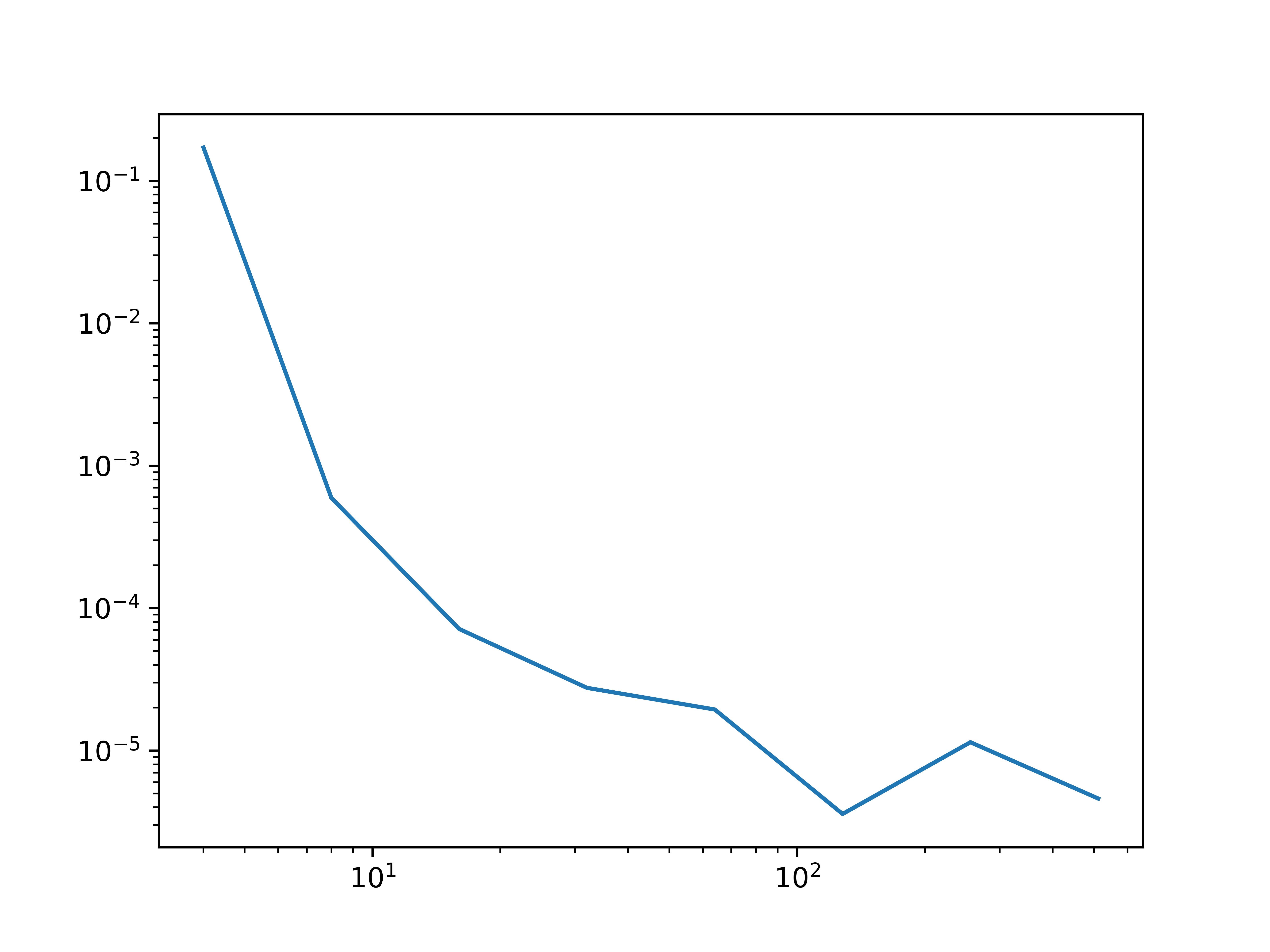}
\caption{The average absolute error as a function of the number of boundary points for the boundary value problem of Example \ref{subsec:radial0}.}\label{BPI_rad_square}
\end{figure}

\begin{figure}[hbt!]
\includegraphics[scale=0.6]{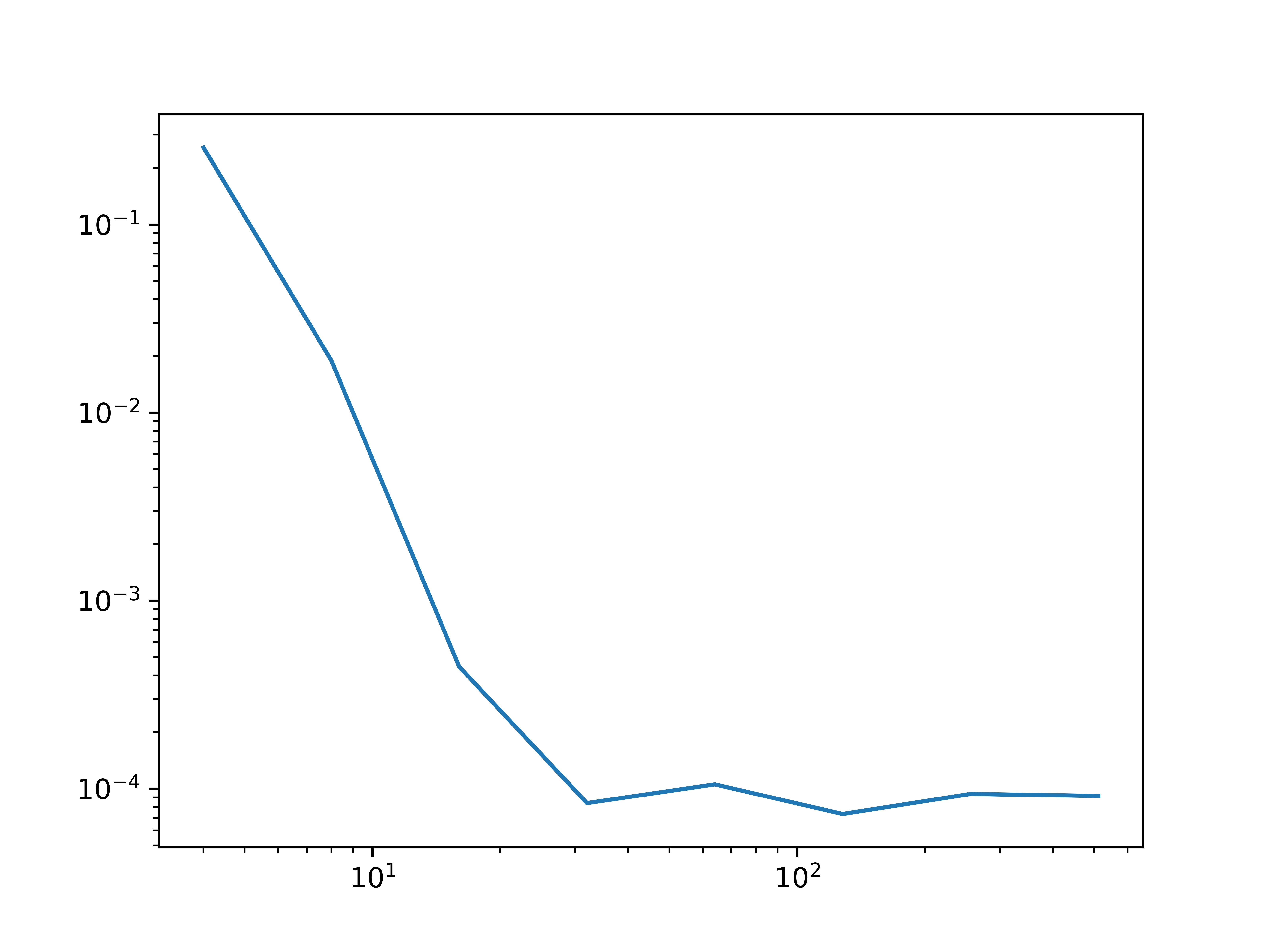}
\caption{The average absolute error as a function of the number of boundary points for the boundary value problem of Example \ref{source_blowup1}.}\label{BPI_blowup}
\end{figure}

\begin{figure}[hbt!]
\includegraphics[scale=0.6]{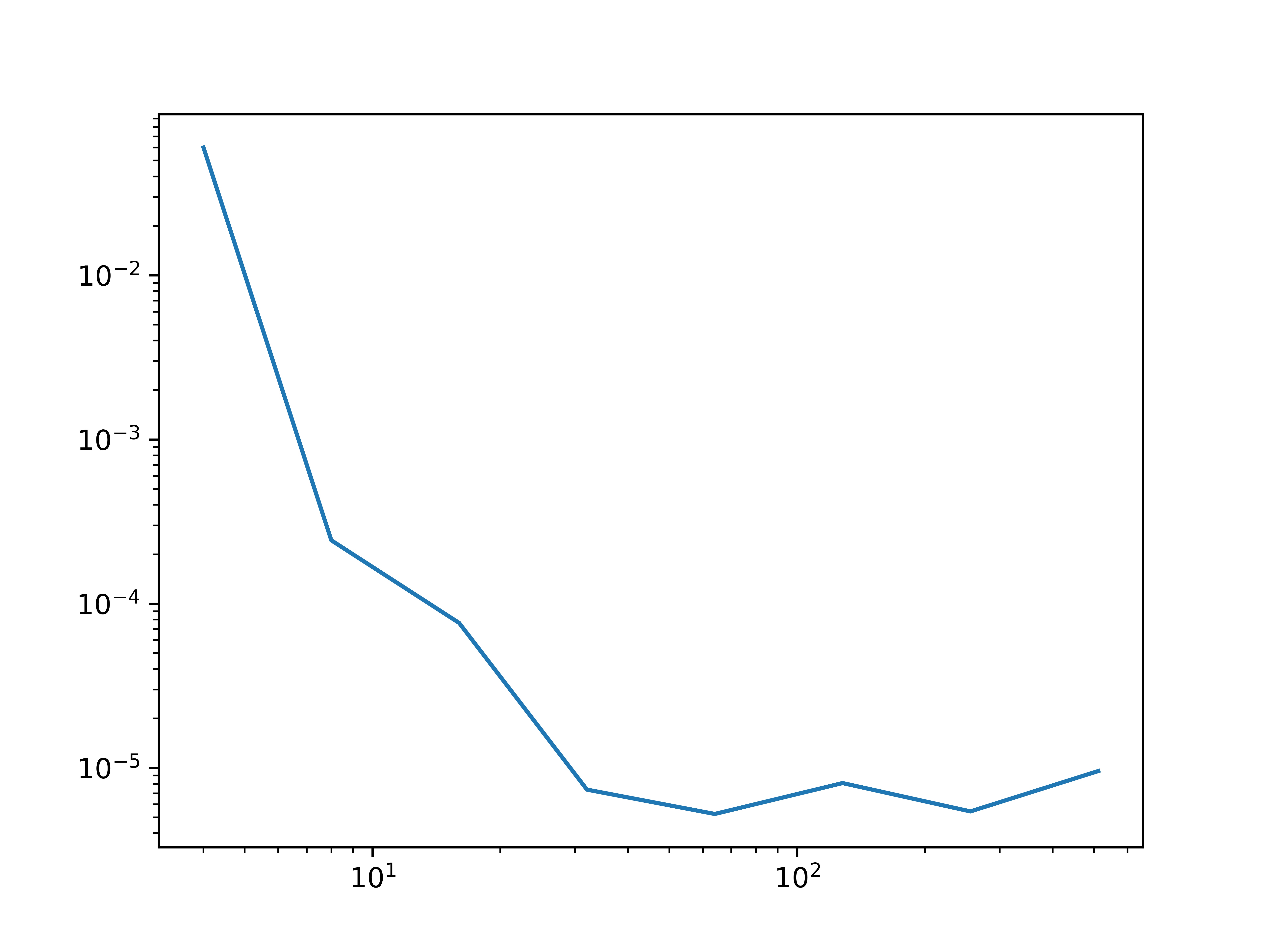}
\caption{The average absolute error as a function of the number of boundary points for the boundary value problem corresponding to the function in \eqref{radial:2dim} on the unit disk.}\label{BPI_rad_disk}
\end{figure}

We fix the number of interior points to be 1000 but vary the number of boundary points.  We train the network using BFGS until convergence or at most 5000 iterations. We start with 4 points on the boundary and double the amount in each computation until no major improvement can be seen. The average error as a function of the number of boundary points for the three examples can be seen in Figures \ref{BPI_rad_square}, \ref{BPI_blowup} and \ref{BPI_rad_disk} respectively.

\subsection{Increasing both interior and boundary points}
We start with a small number of boundary and interior points and double the amount of both types of points in each computation. We train the network using BFGS until convergence or at most 5000 iterations. The results from the three examples can be found in Tables \ref{table:radial_disk_convg}, \ref{table:radial_square_convg} and \ref{table:blowup} respectively.

\begin{table}[h]
\begin{tabular}{|c|c|c|}
boundary points & interior points & average error  \\
\hline
4 & 20  & 0.368 \\
8 & 40  & 1.39e-02 \\
16 & 80  & 1.57e-03 \\
32 & 160  & 5.69e-06 \\
64 & 320  & 4.09e-06 \\
128 & 640 & 6.94e-06 \\
256 & 1280  & 4.49e-06 \\
512 & 2560 & 5.06e-06 \\
1024 & 5120 & 3.05e-06 \\
\end{tabular}\vspace{1mm}\caption{Average error after 5000 iterations for the boundary value problem corresponding to the function in \eqref{radial:2dim} on the unit disk.}\label{table:radial_disk_convg}
\end{table}

\begin{table}[h]
\begin{tabular}{|c|c|c|}
boundary points & interior points & average error  \\
\hline
4 & 10 & 0.158 \\
8 & 20 & 5.83e-04 \\
16 & 40 & 2.29e-04 \\
32 & 80 & 7.46e-06 \\
64 & 160 & 1.08e-05 \\
128 & 320 & 4.50e-06 \\
256 & 640 & 6.46e-06 \\
512 & 1280 & 3.54e-06 
\end{tabular}\vspace{1mm}\caption{Average error after 5000 iterations for the boundary value problem corresponding to the function in \eqref{radial:2dim} on the unit disk.}\label{table:radial_square_convg}
\end{table}

\begin{table}[h]
\begin{tabular}{|c|c|c|}
boundary points & interior points & average error  \\
\hline
4 & 10 & 4.49e-02 \\
8 & 20 & 1.27-03 \\
16 & 40 & 5.06e-04 \\
32 & 80 & 1.98e-04 \\
64 & 160 & 6.39e-05 \\
128 & 320 & 5.36e-05 \\
\end{tabular}\vspace{1mm}\caption{Average error after 5000 iterations for the boundary value problem corresponding to Example \ref{source_blowup1}.}\label{table:blowup}
\end{table}

\vspace{5mm}
The previous examples show that the average absolute error decreases as the number of collocation points increases. When the error becomes sufficiently small, convergence slows down and some oscillation may occur. The main factor preventing better convergence in these examples appears to be the imposed maximum number of iterations, as this number of iterations frequently is reached before the optimization has been terminated successfully in the cases where a large number of collocation points are used. We have verified that one can improve accuracy in these cases simply by increasing the number of iterations. However, naturally this also increases the computation time. When only the number of one type of collocation points is increased while the other type is kept constant, the ratio between the types of points is distorted which could also harm convergence eventually.

Due to the random choice of collocation points and the random initiation of the weights and biases, the outcome of any specific calculation performed above can vary somewhat if it is performed repeatedly with the same number of points, but the general trend is always the same, namely that an increased number of points leads to better accuracy.

\clearpage

\section{Estimating the error}\label{sec:error_est}
In this section we explore how the error in our numerical methods can be estimated. The starting point is the following stability result for solutions to the Dirichlet problem.
\begin{prop}\label{prop:stability} Suppose that $u_1, u_2 \in C(\bar \Omega)$ are convex functions on the bounded domain $\Omega \subset \R^n$ satisfying the Dirichlet problem
\begin{align*}
\left\{
\begin{array}{ll}
\det D^2 u_j = f_j  &\quad \text{ in } \Omega, \\[5pt]
\phantom{\det D^2} u_j = g_j & \quad \text{ on } \partial\Omega,
\end{array}
\right.
\end{align*}
where $f_j \in L^1(\Omega; [0,\infty))$ and $g_j \in C(\partial \Omega)$, $j \in \{1,2\}$. Suppose also that 
\begin{align*}
 \norm{f_1 - f_2}_{L^\infty(\Omega)} < \infty.
\end{align*}
Then
\begin{align}\label{est:stability}
 \norm{u_1 - u_2}_{L^\infty(\Omega)} \leq \sup_{x \in \partial \Omega} |g_1(x) - g_2(x)| + C_n\diam(\Omega) |\Omega|^\frac{1}{n} \norm{f_1 - f_2}_{L^\infty(\Omega)}^\frac{1}{n},
\end{align}
where $C_n$ is a constant only depending on the dimension.
\end{prop}
\begin{proof}{}
 To simplify the notation we write
 \begin{align*}
  \eta &:= \norm{f_1 - f_2}_{L^\infty(\Omega)},
  \\
  \varepsilon &:= \sup_{x \in \partial \Omega} |g_1(x) - g_2(x)|.
 \end{align*}
By the assumptions we have 
\begin{align*}
 M u_2 &\leq f_1 + \eta, \textnormal{ in } \Omega,
 \\
 u_2 &\geq g_1 - \varepsilon, \textnormal{ on } \partial \Omega.
\end{align*}
By the comparison principle of Theorem \ref{ComparisonPrinciple} we have that $u_2 \geq w$ where $w$ is the unique convex solution to the problem
\begin{align*}
\left\{
\begin{array}{ll}
\det D^2 w = f_1 + \eta  &\quad \text{ in } \Omega, \\[5pt]
\phantom{\det D^2} u_j = g_1 - \varepsilon & \quad \text{ on } \partial\Omega.
\end{array}
\right.
\end{align*}
Note that $\tilde w := w + \varepsilon$ is the unique convex solution to the problem
\begin{align*}
\left\{
\begin{array}{ll}
\det D^2 \tilde w = f_1 + \eta  &\quad \text{ in } \Omega, \\[5pt]
\phantom{\det D^2} u_j = g_1  & \quad \text{ on } \partial\Omega.
\end{array}
\right.
\end{align*}
Let $v$ be the unique convex solution to 
\begin{align*}
\left\{
\begin{array}{ll}
\det D^2 v =  \eta  &\quad \text{ in } \Omega, \\[5pt]
\phantom{\det D^2} u_j = 0 & \quad \text{ on } \partial\Omega.
\end{array}
\right.
\end{align*}
By Proposition \ref{prop:sum_of_functions} we have
\begin{align*}
 M(u_1 + v) \geq M u_1 + M v = f_1 + \eta = M \tilde w.
\end{align*}
Moreover, $\tilde w$ and $u_1 + v$ coincide on $\partial \Omega$ so we may again use the comparison principle to conclude that
$\tilde w \geq u_1 + v$ in $\Omega$. Combining the estimates we have
\begin{align}\label{est:u_2}
 u_2 \geq w = \tilde w - \varepsilon \geq u_1 + v - \varepsilon \textnormal{ in } \Omega.
\end{align}
We can now use Theorem \ref{AlexandrovMaxPrinciple} to estimate
\begin{align}\label{est:v}
 |v(x)|^n \leq c_n \diam(\Omega)^{n-1}d(x,\partial \Omega) |M v(\Omega)| \leq \frac{c_n}{2} \diam(\Omega)^n \eta |\Omega|.
\end{align}
Combining \eqref{est:u_2} and \eqref{est:v} we end up with
\begin{align*}
 u_2 \geq u_1 - \varepsilon - C_n\diam(\Omega) |\Omega|^\frac{1}{n} \eta^\frac{1}{n} \textnormal{ in } \Omega,
\end{align*}
where $C_n = (2^{-1} c_n)^\frac1n $. Using the definitions of $\varepsilon$ and $\eta$, and the that one can exchange the roles of $u_1$ and $u_2$ in the previous argument we end up with \eqref{est:stability}.
\end{proof}
The previous result shows that if one can construct a convex function which approximately satisfies the equation and whose boundary values are close to the desired boundary values, then also the function itself is close to the exact solution. This motivates our numerical methods since we are essentially constructing a convex function which approximately satisfies the boundary value problem. 

The previous result can also be used to estimate the error, even if the exact solution is uknown. To see this, suppose that $u$ is the (possibly unknown) solution to
\begin{align}\label{prob:again}
\left\{
\begin{array}{ll}
\det D^2 u = f   &\quad \text{ in } \Omega, \\[5pt]
\phantom{\det D^2} u = g & \quad \text{ on } \partial\Omega.
\end{array}
\right.
\end{align}
Suppose that we have a known smooth convex function $v$ defined on $\bar \Omega$ (for example taking the form of an input convex neural network). Then $v$ satisfies a problem of the same form with the right-hand side $f_v := \det D^2 v$ and the boundary values $g_v:= v|_{\partial \Omega}$. Proposition \ref{prop:stability} states that in order to estimate the error $|u - v|$, it is sufficient to analyze the behavior of the known functions $f$, $g$, $f_v$ and $g_v$. For sufficiently regular $f$ and $g$, this can be done using a finite number of function evaluations. For illustrative purposes we consider in the following the case when $f$ and $g$ are Lipschitz with known Lipschitz constants, although the approach can be adapted also to the case where $f$ and $g$ have some other known moduli of continuity. Specifically we observe that we have the following result.
\begin{lem}\label{lem:finite-evals}
Suppose that $u$ is a solution to \eqref{prob:again} where $f$ and $g$ are Lipschitz. Let $v \in C^3(\bar\Omega)$ be a convex function. Take points $\{x_j\}^M_{j=1} \subset \Omega$, $\{y_k\}^N_{k=1}\subset \partial \Omega$ and numbers $r,s > 0$ such that 
\begin{align*}
 \Omega \subset \cup^M_{j=1} B(x_j,r), \quad \partial \Omega \subset \cup^N_{k=1} B(y_k,s).
\end{align*}
Then
\begin{align}\label{est:finite}
 \norm{u-v}_{L^\infty(\Omega)} &\leq  C_n \diam(\Omega)|\Omega|^\frac1n \big( (\Lip f + \Lip f_v)r + \max_{j\in \{1,\dots,M\}}|f(x_j) - f_v(x_j)|\big)^\frac{1}{n}
 \\
\notag & \quad + (\Lip g + \Lip g_v )s + \max_{k\in \{1,\dots,N\}}|g(y_k) - g_v(y_k)|, 
\end{align}
where $C_n$ is the constant of Proposition \ref{prop:stability}. 
\end{lem}
\begin{proof}{}
 For any point $x\in \Omega$ we find $x_j$ such that $|x-x_j|< r$ and thus
 \begin{align*}
  |f(x) - f_v(x)| &\leq |f(x) - f(x_j)| + |f(x_j) - f_v(x_j)| + |f_v(x_j) - f_v(x)|
  \\
   & \leq (\Lip f )r + |f(x_j) - f_v(x_j)| + (\Lip f_v) r,
 \end{align*}
and hence
\begin{align}\label{est:f-lip}
 \norm{f-f_v}_{L^\infty(\Omega)} \leq (\Lip f + \Lip f_v)r + \max_{j\in \{1,\dots,M\}}|f(x_j) - f_v(x_j)|.
\end{align}
Similarly, we see that 
\begin{align}\label{est:g-lip}
 \sup_{x\in \partial \Omega}|g(y) - g_v(y)| \leq (\Lip g + \Lip g_v )s + \max_{k\in \{1,\dots,N\}}|g(y_k) - g_v(y_k)|.
\end{align}
Combining \eqref{est:f-lip} and \eqref{est:g-lip} with \eqref{est:stability} we end up with \eqref{est:finite}. 
\end{proof}
In order for the previous result to be useful we need to be able to estimate the Lipschitz constants of $f_v$ and $g_v$. Since $g_v = v|_{\partial \Omega}$ it is sufficient to estimate $\nabla v$ in order to obtain a bound for the Lipschitz constant of $v$. Similarly, we want to bound the first order partial derivatives of $f_v$ in order to obtain a bound for $\Lip f_v$. For this purpose, we write $H = D^2 v$ and observe that 
\begin{align*}
 \partial_k f_v = \partial_k (\det H) = \tr ( \adj H \partial_k H) = \sum^n_{i=1} (\adj H \partial_k H)(i,i) = \sum^n_{j=1} \sum^n_{i=1} \adj H(i,j) \partial_k H (j,i),
\end{align*}
where we recall that the adjugate matrix $\adj H$ takes the form
\begin{align*}
 \adj H (i,j) = (-1)^{i+j} \det H^i_j,
\end{align*}
where $H^i_j$ is the $(n-1)\times (n-1)$ matrix obtained by removing the $j$th row and the $i$th column from $H$. We can therefore obtain the rough estimate
\begin{align*}
 |\adj H(i,j)| \leq (n-1)! \Big(\sup_{x\in \Omega, |\alpha|=2} |\partial^\alpha v|\Big)^{n-1}.
\end{align*}
Recalling that $\partial_k H$ is a matrix consisting of third order derivatives of $v$ we thus have
\begin{align*}
 |\partial_k f_v| \leq n(n!) \Big(\sup_{x\in \Omega, |\alpha|=2} |\partial^\alpha v|\Big)^{n-1} \sup_{x\in \Omega, |\beta|=3} |\partial^\beta v|,
\end{align*}
and hence
\begin{align}\label{est:nabla-fv}
 |\nabla f_v| \leq \sqrt n n(n!) \Big(\sup_{x\in \Omega, |\alpha|=2} |\partial^\alpha v|\Big)^{n-1} \sup_{x\in \Omega, |\beta|=3} |\partial^\beta v|.
\end{align}

From the previous calculations it follows that in order to bound the Lipschitz constants of $g_v$ and $f_v$ it is sufficient to have bounds for the partial derivatives of $v$ up to order three. These observations are valid for any smooth convex $v$. In the following we will prove bounds for the derivatives in the case that $v$ is an input convex neural network with softplus as the activation function, i.e. the type of networks that we have mainly focused on throughout the paper. We are thus considering a function $v$ of the form
\begin{align}\label{input-conv-special-case}
z_1 &= S(L_0 x + b_0),
\\
\notag z_{j+1} &= S(W_j z_j + L_j x + b_j), \textrm{ for } j \in \{1, \dots, N-1 \},
\\
\notag z_{N+1} &= W_N z_N + L_N x + b_N
\\
\notag v(x) &= z_{N+1},
\end{align}
where  $S(x) = \ln(1+e^x)$. Note that we have no activation function on the last layer, just as in our numerical examples. In order to facilitate the formulation of our estimates we introduce some ulterior notation. By $|L_j|$ intend the matrix obtained by taking the absolute value of all entries in $L_j$. Furthermore, we define the matrices
\begin{align*}
 M_0 &:= |L_0|, \quad M_1 := |L_1| + W_1 |L_0|, \quad M_2 = |L_2| + W_2|L_1| + W_2 W_1|L_0|,
 \\
 M_p &:= |L_p| + W_p M_{p-1} = |L_p| + W_p |L_{p-1}| + W_p W_{p-1} |L_{p-2}| + \dots + W_p\dots W_1 |L_0|.
\end{align*}
We remark that since every matrix $W_j$ has nonnegative entries, the same is true for the matrices $M_j$. Now we can state the following.
\begin{lem}\label{lem:derestimates}
 Let $v$ be a neural network of the form \eqref{input-conv-special-case}. Then for $k,l,r \in \{1, \dots , n\}$,
 \begin{align}
 \label{der-1-v} |\partial_k v| &\leq M_N(k),
  \\
 \label{der-2-v} |\partial_k \partial_l v| &\leq \sum^N_{j=1} \sum^{m_j}_{i=1} (W_N\dots W_j)(i) M_{j-1}(i,k) M_{j-1}(i,l),
  \\
 \label{der-3-v} |\partial_k \partial_l \partial_r v| &\leq \sum^N_{q=1} \sum^{m_q}_{j=1} (W_N\dots W_q)(j) M_{q-1}(j,k) M_{q-1}(j,l) M_{q-1}(j,r)
 \\
 \notag  + \perm_{k,l,r} \sum^{N-1}_{q=1}  \sum^{N-1}_{\nu=q} &\sum^{m_q}_{i=1} \sum^{m_{\nu+1}}_{j=1} (W_N \dots W_{\nu +1})(j)(W_\nu \dots W_q)(j,i) M_{q-1}(i,k) M_{q-1}(i,l) M_\nu(j,r).
 \end{align}
 where we use the abbreviated notation $M_N(k) = M_N(1,k)$ since $M_N$ has only one row. Likewise, we suppress the row number for matrices of the form $W_N \dots W_j$ since again, these have only one row. With $m_j$ we intend the number of columns of $W_j$, which is also the number of rows of $W_{j-1}$ and $M_{j-1}$. On the last row, $\perm$ refers to the sum of the three terms obtained by cyclically permuting $k$, $l$ and $r$. 
\end{lem}
\begin{proof}{}
 For $p \in \{1,\dots, N\}$ we denote 
 \begin{align*}
  A^p = W_p z_p + L_p x + b_p.
 \end{align*}
We remark that $A^p$ in general is a vector valued quantity (except for $p=N$ since the output of the network is one-dimensional). We denote the $j$th component of $A^p$ by $A^p_j$. Using the fact that 
\begin{align*}
 |S'| \leq 1, \quad |S''| \leq 1, \quad |S'''| \leq 1
\end{align*}
and the chain rule, we obtain for $k,l,r \in \{1,\dots,n\}$ the estimates
\begin{align}
 \label{der-1-A1} |\partial_k A^1_s| &\leq M_1(s,k),
 \\
 \label{der-2-A1} |\partial_l \partial_k A^1_s| &\leq \sum^{m_1}_{j_1=1} W_1(s, j_1) M_0(j_1, k) M_0(j_1, l),
 \\
 \label{der-3-A1} |\partial_r \partial_l \partial_k A^1_s| &\leq \sum^{m_1}_{j_1=1} W_1(s, j_1) M_0(j_1, k) M_0(j_1, l) M_0(j_1, r),
\end{align}
where we have introduced the notation $m_p$ for the number of columns of $W_p$. Similarly, since 
\begin{align*}
 A^{p} = W_{p} S(A^{p-1}) + L_{p}x + b_{p}
\end{align*}
we have for $p\in \{2,\dots N\}$ the recursive estimates
\begin{align}
 \label{der-1-recurs} |\partial_k A^p_s| &\leq \sum^{m_p}_{j=1} W_p(s,j)|\partial_k A^{p-1}_j| + |L_p|(s,k),
 \\
 \label{der-2-recurs} |\partial_k \partial_l A^p_s | &\leq \sum^{m_p}_{j=1} W_p(s,j)\big( |\partial_k A^{p-1}_j| |\partial_l A^{p-1}_j| + |\partial_k \partial_l A^{p-1}_j |\big),
 \\
 \label{der-3-recurs} |\partial_k \partial_l \partial_r A^p_s| &\leq \sum^{m_p}_{j=1} W_p(s,j)\Big[ |\partial_k A^{p-1}_j| |\partial_l A^{p-1}_j| |\partial_k A^{p-1}_j| 
 \\
 \notag & \qquad \qquad \qquad \quad + \perm_{k,l,r} \big(|\partial_k A^{p-1}_j| |\partial_l \partial_r A^{p-1}_j|\big) + |\partial_k \partial_l \partial_r A^{p-1}_j| \Big],
\end{align}
where the operator $\perm$ indicates the sum of the three terms obtained by cyclically permuting $k$, $l$ and $r$ inside the expression. Using \eqref{der-1-A1} and \eqref{der-1-recurs} we see by induction that 
\begin{align}\label{der-1-final}
 |\partial_k A^p_s| \leq M_p(s,k).
\end{align}
Combining \eqref{der-2-A1}, \eqref{der-2-recurs} and \eqref{der-1-final} we can verify by induction that 
\begin{align}\label{der-2-final}
 |\partial_k \partial_l A^p_s| \leq \sum^p_{j=1} \sum^{m_j}_{i=1} (W_p\dots W_j)(s,i) M_{j-1}(i,k) M_{j-1}(i,l).
\end{align}
Finally, using \eqref{der-3-A1}, \eqref{der-3-recurs}, \eqref{der-1-final} and \eqref{der-2-final} we can prove using induction that for $p\geq 2$,
\begin{align}\label{der-3-final}
 |\partial_k \partial_l \partial_r A^p_s| &\leq \sum^p_{q=1} \sum^{m_q}_{j=1} (W_p\dots W_q)(s,j) M_{q-1}(j,k) M_{q-1}(j,l) M_{q-1}(j,r)
 \\
 \notag  + \perm_{k,l,r} \sum^{p-1}_{q=1}  \sum^{p-1}_{\nu=q} &\sum^{m_q}_{i=1} \sum^{m_{\nu+1}}_{j=1} (W_p \dots W_{\nu +1})(s,j)(W_\nu \dots W_q)(j,i) M_{q-1}(i,k) M_{q-1}(i,l) M_\nu(j,r).
\end{align}
Since $v=A^N$ which is one-dimensional (i.e. $s=1$ is the only possible choice) we obtain the desired estimates from \eqref{der-1-final}, \eqref{der-2-final} and \eqref{der-3-final} in the special case $p=N$.
\end{proof}

Combining \eqref{der-2-v} and \eqref{der-3-v} with \eqref{est:nabla-fv} it is possible to obtain an estimate for the Lipschitz constant of $f_v$. Using \eqref{der-1-v} we also get an estimate for the Lipschitz constant of $g_v = v|_{\partial\Omega}$. Using these bounds together with Lemma \ref{lem:finite-evals} we can get an upper bound for the error $\norm{u-v}_{L^\infty(\Omega)}$ when $v$ is a neural network of the form \eqref{input-conv-special-case}, by performing a finite number of function evaluations of the known functions $f,g,f_v,g_v$.

\section{Summary and conclusions}
We have discussed and illustrated neural networks as an ansatz for solving the Dirichlet problem for the Monge-Amp\`ere equation. We have devised two approaches: a first approach based on a modified loss function
constructed by adding a term to the loss function which penalizes non-convexity, and a second approach based on input convex neural networks. The first approach is applicable in two dimensions while, in theory, the second approach applies in all dimensions. The methods, and in particular the method based on input convex neural networks, are illustrated in a set of examples covering several dimensions of complexity.

Our numerical illustrations show that deep input convex neural networks provide a good ansatz for solving the Dirichlet problem for the Monge-Amp\`ere equation. The convexity of the network with respect to the input variable guarantees that the unique convex solution is approximated. In our comparison this network architecture turned out to give a far more accurate result than an ansatz of the type used in \cite{BeNy} which, naturally, is more applicable for linear problems. For smooth source functions and boundary data the accuracy of the  method based on input convex neural networks is very high, especially compared with previous neural network based works such as \cite{BeNy}.

Singularities of the source function lying on the boundary can decrease accuracy and efficiency, but this can be remedied by choosing more collocation points near the singularity, as was seen in Examples \ref{source_blowup1} and \ref{surface_of_sphere}. Discontinuities of the source function also have a negative effect on the accuracy, as we have seen in Example \ref{discont_source}. Increasing the number of points near the discontinuities prevents efficient training in this situation. Instead, the accuracy can be improved by increasing the number of points throughout the domain. 

The method involving input convex networks can handle nontrivial domains and is stable under the addition of noise in the source term. We obtained satisfactory results also for three- and four-dimensional domains, and in principle the method should work in any dimension. As we have seen, in practice there are still some challenges remaining when the dimension grows too large. In this area there is room for further research.

As the results of Section \ref{sec:numconvg} show, the methods provide surprisingly good results even if a small number of collocation points is used. Increasing the number of points increases the accuracy. Due to TensorFlow being optimized for calculating many derivatives simultaenously, computation time does not increase much per iteration even though the number of points is increased.  However more points often require more iterations of the BFGS algorigthm in order for the optimization to finish, which increases the overall computation time.

In Section \ref{sec:error_est} we saw that it is possible to estimate the error using a finite number of evaluation of known functions, even in the case that the exact solution is unknown, at least if the right-hand side and the boundary values are sufficiently regular. 

Concerning future research there is plenty of interesting work to be done. First, it is interesting to conduct numerical studies for more general equations of the form
$$\det D^2u = f(x,\,u,\,\nabla u),$$
and to focus, in particular, on the prescribed Gaussian curvature problem and the Monge problem in transport theory. For transport problems, the ``boundary'' condition is somewhat more complicated than our case as it involves the behavior of the gradient on the whole domain. However, as observed by \cite{Po2}, \cite{TruWa} and \cite{Ur}, it is sufficient to require that the gradient maps boundary points to boundary points, and this simplification should allow us to construct a suitable loss function also in this case. In general it is interesting to investigate the effect of using deeper compared to shallow neural networks. Estimating and minimizing the computational time is another area where more research is needed. 
Improving the error estimates of Section \ref{sec:error_est} could be another interesting topic. For example, it would be useful to have better bounds for the Lipschitz constants, reducing the number of function evaluations needed to estimate the error.

\appendix
\section{Additional background for the Monge-Amp\`ere equation}\label{app:MA}
Here we state some additional results regarding the Monge-Amp\`ere measure and provide a sketch of the proof for Theorem \ref{thmdp} for strictly convex $\Omega$.
We start with the Alexandrov maximum principle, see \cite[Theorem 2.8]{F}.
\begin{thm}\label{AlexandrovMaxPrinciple}
 Assume that  $u$ is convex on a bounded convex domain $\Omega\subset\mathbb R^n$ and that $u|_{\partial \Omega} = 0$. Then
 $$|u(x)|^n \leq c d(x,\partial \Omega)(\diam(\Omega))^{n-1}Mu(\Omega)$$
 for a constant $c$ which only depends on $n$.
\end{thm}

Theorem \ref{AlexandrovMaxPrinciple} states  that functions with bounded Monge-Amp\`{e}re mass have a $C^{1/n}$ modulus of continuity near the boundary of a sublevel set. 
The following result is found in \cite[Lemma 2.9]{F}.
\begin{prop}\label{prop:sum_of_functions}
 Assume that  $u$ and $v$ are convex functions on a bounded convex domain $\Omega\subset\mathbb R^n$. Then
 \begin{align*}
  M(u+v) \geq M u + M v.
 \end{align*}
\end{prop}
The set of Alexandrov solutions turns out to be closed under uniform convergence, see \cite[Proposition 2.6]{F}.
\begin{prop}\label{Closedness}
If $u_k$ converge uniformly to $u$ in $\Omega \subset \mathbb{R}^n$, then $Mu_k$ converges weakly to $Mu$ in $\Omega$.
\end{prop}

Theorem \ref{AlexandrovMaxPrinciple} and Proposition \ref{Closedness} imply the following compactness result for solutions with zero boundary data, see \cite[Corollary 2.12]{F} for a more general statement.

\begin{prop}\label{Compactness} For a bounded convex domain $\Omega\subset\mathbb R^n$, and a finite constant $c$, consider the collection of functions $$\mathcal{A} = \{v:\ v \text{ convex on } \Omega,\, v = 0\mbox{ on }\partial\Omega, \, Mv(\Omega) \leq c\}.$$
Then $\mathcal{A}$ compact in the sense that any sequence in $\mathcal{A}$ has a locally uniformly convergent subsequence in $\Omega$ whose Monge-Amp\`{e}re measures converge weakly to the Monge-Amp\`{e}re measure of the limit.
\end{prop}

Using \ref{ComparisonPrinciple} and Proposition \ref{Compactness} we can now sketch the proof of existence and uniqueness of solutions to the Dirichlet problem in the case of strictly convex $\Omega$.

\vspace{2mm}
\noindent \textbf{Sketch of the proof of Theorem \ref{thmdp} for solutions vanishing on the boundary.}
The  uniqueness part of the theorem is a consequence of the comparison principle stated in Theorem \ref{ComparisonPrinciple}. To give a flavour
of the proof of existence we here only consider the case when $\Omega$ is a polyhedron and $g = 0$ following \cite[Theorem 1]{Mo}. See also \cite[Theorem 2.13]{F} where a somewhat different approach is used to directly treat general convex $\Omega$ with $g=0$. As $\mu$ can be approximated in the weak* topology by finite sums of Dirac masses, $\sum_{i = 1}^N \alpha_i \delta_{x_i}$, Proposition \ref{Compactness} implies that it suffices to consider this case.

Let $\mathcal{F}$ be the family of convex polyhedral graphs $P$ in $\mathbb{R}^{n+1}$ that contain $(\partial \Omega,\, 0) \subset \mathbb{R}^n \times \mathbb{R}$, with remaining vertices that project to a subset of $\{x_i\}_{i = 1}^N$. Let
$\mathcal{F}' \subset \mathcal{F}$ consist of those $P$ satisfying $MP \leq \sum_{i = 1}^N \alpha_i \delta_{x_i}$. The family $\mathcal{F}'$ is non-empty and compact in the sense of Proposition \ref{Compactness}. For $P \in \mathcal{F}$ we let $\phi(P) = \sum_{i = 1}^N P(x_i)$. The functional $\phi$ is bounded below on $\mathcal{F}'$ by Theorem \ref{AlexandrovMaxPrinciple}. By compactness
there exists a minimizer $u$ of $\phi$ in $\mathcal{F}'$. We want to conclude that $u$ solves $Mu = \sum_{i = 1}^N \alpha_i \delta_{x_i}$. Assume not, then  $Mu(\{x_k\}) < \alpha_k$ for some $k\in\{1,...,N\}$. By moving the vertex
$(x_k,\, u(x_k))$ slightly downwards and taking the convex hull of this point with the remaining vertices, we then obtain another function in $\mathcal{F}'$ that is smaller than $u$. This a contradiction and hence we can conclude that $Mu = \sum_{i=1}^N \alpha_i \delta_{x_i}$.

\vspace{2mm}
The case when $\Omega$ is a polyhedron, and $g$ is affine on each face of $\partial \Omega$, can be treated similarly with $\mathcal{F},\, \mathcal{F}'$ consisting of convex polyhedral graphs with vertices over $\{x_i\}_{i = 1}^N$ and
$g$ as boundary data. To prove  that $\mathcal{F}'$ is non-empty in this case, one instead uses the convex hull of the graph of $g$ in $\mathbb{R}^{n+1}$.

For the general case, one can approximate $\Omega$ with the convex hulls of finite subsets of $\partial \Omega$ with finer and finer mesh and one needs to approximate $g$ with data which are affine on the faces
of these polyhedra. One then has to solve these approximating problems and take a limit. The strict convexity of $\partial \Omega$ is used in the last step. It guarantees that for any subset $\{y_i\}_{i = 1}^M$ of $\partial \Omega$, each $y_k$ is a vertex of the convex hull of $\{y_i\}_{i = 1}^M$. We refer the reader to \cite{CY}, for details. See also the approach in \cite[Theorem 2.14]{F} where the approximation with convex polyhedra is avoided.

Note that when $g$ is linear, we don't require strict convexity of $\partial \Omega$.
The strict convexity is necessary for general $g$ since no convex function can continuously attain e.g. the boundary data $-|x|^2$ when $\partial \Omega$ has flat pieces.

\section{The approximation problem}\label{app:approximation}
Here we provide an overview of results related to the approximation capabilities of (deep) neural networks. We note that there is an abundance of different neural network representations one could attempt to use as an
ansatz for the solution $u$ of the Monge-Amp\`ere equation. Recall that $u:\Omega\subset\mathbb R^n\to\mathbb R$. Several of the
results stated in the literature concern the approximation by neural networks considering $[0,1]^n$ as the underlying domain. These results are applicable in more general domains $\Omega$ whenever a (regular) map $[0,1]^n\to \bar \Omega$ exists. We first recall a version of the Kolmogorov Superposition Theorem, see \cite{Lo}, which states that any continuous function defined on $[0,1]^n$, can be represented (exactly), for an  appropriate activation function, as a neural network. For a recent and informative 
account of the Kolmogorov Superposition Theorem we refer to \cite{Liuu}.
\begin{thm}\label{thm:kol}
    There exist $n$ constants $\{\lambda_j\}$, $\lambda_j>0$, $\sum_{j=1}^n \lambda_j\leq 1$ and $2n+1$ strictly increasing continuous functions $\{\phi_i\}$, each of which mapping $[0,1]$ to itself, such that
    if $u\in C([0,1]^n)$, then $u$ can be represented in the form
    \begin{equation}\label{equ:super}
       u(x)=u(x_1, \ldots, x_n) = \sum_{i=1}^{2n+1} h\left( \sum_{j=1}^n \lambda_j \phi_i(x_j) \right)
    \end{equation}
    for some $h\in C([0,1])$ depending on $u$.
\end{thm}

While Theorem \ref{thm:kol} stresses the power of (general) neural networks to represent functions, it adds little in practice as we usually want to design the neural network without a priori knowledge of $u$.  In the following we first discuss the approximation capabilities of networks with one and two hidden layers before giving some references
to more recent results concerning the approximation capabilities of (deep) neural networks.

The approximation properties of  neural networks with one hidden layer is rather well understood. As previously mentioned,  neural networks with one hidden layer can approximate any continuous function to arbitrary accuracy by having a sufficient amount of parameters \cite{Hornik1990551}. We here formulate another result which can found in \cite{mhaskar1996neural}. Given $1\leq p\leq \infty$ and $r\geq 1$, we define  $W^p_{r,n}:=W^p_{r}((-1,1)^n)$ as the space of functions $u$ on $(-1,1)^n$ for which
 \begin{eqnarray*}
     \|u\|_{W^p_{r,n}} := \sum_{0\leq k\leq r} \|\nabla^{{k}}u\|_{p}<\infty,
 \end{eqnarray*}
 where $\|\cdot\|_p$ denotes the $L^p$ norm on $(-1,1)^n$.
\begin{thm}\label{thm:1d}
     Let $1\leq n'\leq n$ and let $\phi:\mathbb R^{n'}\rightarrow\mathbb R$ be a function which is infinitely  differentiable in some open sphere in $\mathbb R^{n'}$. Assume also that there exists ${b}$ in this sphere such that
     \begin{equation*}
         \nabla^k\phi(b)\neq 0\mbox{ for all }k\geq 0.
     \end{equation*}
     Then there exist $n'\times n$ matrices $\{W_j\}_{j=1}^n$ with the property that there exist, given $u\in W^p_{r,n}$, coefficients $a_j(u)$ such that
     \begin{equation*}
         \|u(\cdot)-\sum_{j=1}^k a_j(u) \phi(W_j(\cdot)+{b})\|_p \leq ck^{-r/n}\|u\|_{W^p_{r,n}}.
     \end{equation*}
 \end{thm}

 Theorem \ref{thm:1d} applies when $n'=1$, and with
 $\phi=\sigma$ being the sigmoid activation function. Note that in the case $r=1$, then Theorem \ref{thm:1d} produces  a neural network with one hidden layer and with hidden size equal to $n$. To achieve a precision
 in the approximation of $u\in W^p_{1,n}$ with a
 predetermined error $\varepsilon>0$, we need
 \begin{eqnarray*}\label{equ:exp}
     k^{-1/n} = \mathcal{O}(\varepsilon) \Rightarrow k = \mathcal{O}(\varepsilon^{-n}).
 \end{eqnarray*}
 This observation indicates that neural networks with one hidden layer suffer from the curse of dimensionality in the sense that the number of required neurons in the hidden layer grows exponentially with the input dimension $n$. A  possible way to overcome this, motivated by  the Kolmogorov Superposition Theorem stated in Theorem \ref{thm:kol}, is to consider multi-layer neural networks.  Considering neural networks with two hidden layers, in \cite{maiorov1999lower} it is proved that if the  number of units in the hidden layers are $6n+3$ and $3n$, respectively, then these networks can approximate any function to arbitrary precision.
\begin{thm}\label{thm:mai}
    There exists an analytical, strictly increasing activation function $\sigma$ such that if $u\in C([0,1]^n)$ and $\varepsilon>0$, then there exist constants $d_i$, $c_{ij}$, $\theta_{ij}$, $\gamma_i$ and vectors $\mathbf{w}^{ij}\in \mathbb R^n$ for which
    \[\left| {u(x) - \sum\limits_{i = 1}^{6n + 3} {{d_i}\sigma \left( {\sum\limits_{j = 1}^{3n} {{c_{ij}}\sigma ({{\mathbf{w}}^{ij}} \cdot x - {\theta _{ij}})}  - {\gamma _i}} \right)} } \right| < \varepsilon \]
    for all $x\in [0,1]^n$.
\end{thm}

Note that as in Theorem \ref{thm:kol}, Theorem \ref{thm:mai} only implies the approximation result for appropriate activation functions. As it turns out, these activation functions can be quite pathological to achieve the desired accuracy and its is unclear if the activation functions used in practice, for example the sigmoid activation function, can be used to achieve the approximation result. However, stronger results can be achieved if more is known about the function $u$. Indeed, given $u\in C([0,1]^n)$,  consider the superposition decomposition as in \eqref{equ:super}. Consider the family of the functions $u$ where the corresponding $h\in W_{1,1}^p$ and $\phi_i\in W_{1,1}^p$. Then by Theorem \ref{thm:1d}, there exist constants ${{w_{ri}}}$, ${{a_{ri}}}$, ${{b_{ri}}}$ and a constant $C>0$ independent of $r$, such that
\begin{eqnarray*}\label{equ:gt}
\left| {h(t) - \sum\limits_{i = 1}^r {{w_{ri}}\sigma ({a_{ri}}t + {b_{ri}})} } \right| \leqslant \frac{{C||h|{|_{W_{1,1}^p}}}}{r}.
\end{eqnarray*}
We say that $u$ satisfies regularity bounds with constant $\tilde C$ if, for all $r$,
\begin{eqnarray*}
    |w_{ri}|\leq \tilde C, |a_{ri}|\leq \tilde C\qquad i = 1,2,\ldots, r.
\end{eqnarray*}
For  a function $u$ satisfying regularity bounds with constant $\tilde C$ one can derive an explicit error bound for the approximation by a neural network with two hidden layers and with sigmoid activation functions.  For the proof of the following theorem we refer
to \cite{2018arXiv181208883X}.
 \begin{thm}\label{thm:kai}
     Let $\sigma$ be the sigmoid activation
     $( \sigma(x)={1}/{(1+e^{-x})})$. Let  $u\in C([0,1]^n)$ be any function
     satisfying regularity bounds with constant $\tilde C$ and let
     $\varepsilon>0$. Then there exists a  neural network with two hidden layers, with this particular $\sigma$ as the activation function, with hidden units $s=\mathcal{O}\left( \frac{n^2}{\varepsilon^2} \right)$~(the first layer) and $r=\mathcal{O}\left( \frac{n}{\varepsilon} \right)$~(the second layer), and constants $d_i$, $c_{ij}$, $\theta_{ij}$, $\gamma_i$ and vectors $\mathbf{w}^{ij}\in \mathbb R^n$ such that the conclusion of Theorem
     \ref{thm:mai} holds.
     \end{thm}

 Theorem \ref{thm:1d} and Theorem \ref{thm:kai} shed some light on the relevance and potential advantage of using  neural networks with two hidden layers compared to neural networks with one hidden layer when approximating functions. In general, rigorous yet practically applicable theorems concerning approximation using deep neural networks are still lacking. However, there is an emerging literature on the topic and we refer the interested reader to \cite{we1,we2,we3} and the references therein for surveys and introductions to the subject and its context.


\begin{thebibliography}{99}
\bibitem{TensorFlow} M. Abadi et al., \emph{TensorFlow: Large-scale machine learning on heterogeneous systems}, Software available from tensorflow.org.




    \bibitem{AmXuKo} B.~Amos, L.~Xu, J.~Z.~Kolter: \emph{Input Convex Neural Networks}, Proceedings of the 34th International Conference on Machine Learning, PMLR 70:146--155, 2017.

\bibitem{BarS}
G. Barles,  P.E. Souganidis: \emph{Convergence of approximation schemes for fully nonlinear second order equations}, Asymptotic Analysis, 4(3):271--283, 1991.






\bibitem{BeFroeOb0}
J.-D. Benamou, B.D. Froese, and A.M. Oberman.
\newblock Two numerical methods for the elliptic {M}onge-{A}mp\`ere equation.
\newblock {\em ESAIM: Mathematical Modelling and Numerical Analysis},
  44(4):737--758, 2010.


\bibitem{BeNy} J. Berg and K. Nystr{\"o}m: \emph{A unified deep artificial neural network approach to partial differential equations in complex geometries}, Neurocomputing, 317, 28-41, 2018.






    \bibitem{Br} Y. Brenier: \emph{Polar factorization and monotone rearrangement of vector-valued functions}, {Comm. Pure Appl. Math.}, {44}(4), 375--417, 1991.

        \bibitem{brenner-et-al:2011-1}
S.C. Brenner, T.~Gudi, M.~Neilan, L.Y. Sung: \emph{{C}$^0$ penalty methods for the fully nonlinear {M}onge-{A}mp\`ere
  equation},  {Mathematics of Computation}, 80, 1979--1995, 2011.

\bibitem{BriHaPla} K.~Brix, Y.~Hafizogullari, A.~Platen: \emph{Solving the Monge-Amp\`ere equations for the inverse reflector problem}, Mathematical Models and Methods in Applied Sciences, 25 (05), 803-837, 2015.

\bibitem{BriHaPla2} K.~Brix, Y.~Hafizogullari,  A.~Platen: \emph{Designing illumination lenses and mirrors by the numerical solution of Monge-Amp\`ere equations}, Journal of the Optical Society of America A, 32 (11),  2227--2236, 2015.

    \bibitem{Browne}
P.~Browne, C.~Budd, C.~Piccolo, M.~Cullen: \emph{Fast three dimensional r-adaptive mesh
redistribution}, Journal of Computational Physics, 275:174--196, 2014

\bibitem{Budd}
C.~Budd, J.~Williams: \emph{Moving mesh generation using the parabolic Monge-Amp\`ere Equation},
SIAM Journal on Scientific Computing, 31(5):3438--3465, 2009.


\bibitem{BS}
K.~B\"ohmer, R.~Schaback: \emph{A Meshfree Method for Solving the Monge-Amp\`ere Equation}, Numerical Algorithms, 82, 539--551, 2019.

\bibitem{CaGloGo} A.~Caboussat, R.~Glowinski, G.~Gourzoulidis: \emph{A Least-Squares/Relaxation Method for the Numerical Solution of the Three-Dimensional Elliptic Monge–Amp\`ere Equation}, Journal of Scientific Computing, 2018.


\bibitem{CaGloSo} A.~Caboussat, R.~Glowinski, D.~C.~Sorensen: \emph{A least-squares method for the numerical solution of the Dirichlet problem for the elliptic Monge-Amp\`ere equation in dimension two}, ESAIM Control Optimisation and Calculus of Variations, 19(3), 780--810, 2013.

\bibitem{CC} L.~Caffarelli,  X.~Cabr\'{e}: {\it Fully nonlinear elliptic equations}, {Amer. Math. Soc. Colloq. Publ.} 43,  American Mathematical Society, 1995.

\bibitem{CaKoOl}L.~Caffarelli, S.~A.~Kochengin, V.~I.~Oliker: \emph{On the Numerical Solution of the Problem of Reflector Design with Given Far-Field Scattering Data},  Contemporary Mathematics, 226, 2000.

\bibitem{Ca} E.~Calabi: \emph{Complete affine hyperspheres} I. Symposia Mathematica, Vol. X, (Convegno di Geometria Differenziale, INDAM, Rome, 1971), 19-38, Academic Press, London, 1972.

\bibitem{CheY} S.-Y.~Cheng, S.-T.~Yau: \emph{Complete affine hypersurfaces. I. The completeness of affine metrics}, Comm. Pure Appl. Math. 39, no. 6, 839-866, 1986.

    \bibitem{CY}  S.-Y.~Cheng, S.-T.~Yau: \emph{On the regularity of the Monge-Amp\`ere equation $\det \partial^2u/\partial x_i\partial x_j=F(x,u)$}, Comm. Pure Appl. Math., {30}(1), 41-68, 1977.



\bibitem{DG}
E. Dean, R. Glowinski: \emph{Numerical methods for fully nonlinear elliptic equations of the
Monge-Amp\`ere type}, {Computer Methods in Applied Mechanics and Engineering}, 195(13-16), 1344--1386, 2006.



\bibitem{Delz}
G.Delzanno, L. Chacfon, J. Finn, Y. Chung, G. Lapenta: \emph{An optimal robust equidistribution
method for two-dimensional grid adaptation based on Monge-Kantorovich optimization},  Journal
of Computational Physics, 227(23), 9841--9864, 2008.

\bibitem{FJ}
X. Feng, M. Jensen: \emph{Convergent semi-Lagrangian methods for the Monge-Amp\`ere equation
on unstructured grids}, SIAM J. Numer. Anal., 55(2), 691--712, 2017.


\bibitem{feng-neilan:2009-1}
X.~Feng,  M.~Neilan: \emph{Mixed finite element methods for the fully nonlinear
  {M}onge-{A}mp\`ere equation based on the vanishing moment method}, SIAM J. Numer. Anal., 47, 1226--1250, 2009.



\bibitem{F} A. Figalli: {\it The Monge-Amp\`{e}re Equation and its Applications}. Z\"{u}rich Lectures in Advanced Mathematics. European Mathematical Society (EMS), Z\"{u}rich, 2017.



    \bibitem{Fro}
B.D Froese: \emph{Numerical Methods for the Elliptic Monge-Amp\`ere Equation and Optimal Transport},
Phd, Simon Fraser University, 2012.

\bibitem{froese-oberman:2011-1}
B.D. Froese, A.M. Oberman: \emph{Convergent finite difference solvers for viscosity solutions of the
  elliptic {M}onge-{A}mp\`ere equation in dimensions two and higher}, {SIAM Journal on Numerical Analysis}, 49(4), 1692--1714, 2011.





\bibitem{froese-oberman:2013-1}
B.D. Froese, A.M. Oberman: \emph{Convergent filtered schemes for the {M}onge--{A}mp\`ere partial
  differential equation}, {SIAM J. Numer. Anal.}, 51(1), 423--444, 2013.

    \bibitem{difficultdeep}
    X. Glorot, Y. Bengio: \emph{Understanding the difficulty of training deep feedforward neural
networks}, In Proceedings of the Thirteenth International Conference on Artificial Intelligence and Statistics, 249--256. PMLR, 2010.

\bibitem{Goodfellow-et-al-2016}
I. Goodfellow, Y. Bengio, A. Courville: \emph{Deep Learning}, {MIT Press}, 2016.

\bibitem{Gut} C.~Guti\'errez: {\it The Monge-Amp\`{e}re Equation}, Progress in Nonlinear Differential Equations and their Applications 44, Birkh\"{a}user Boston, Inc., Boston, MA, 2001.

\bibitem{Ha} D.~Hartenstine: \emph{The Dirichlet problem for the Monge-Amp\`ere equation in convex (but not strictly convex) domains}, Electronic Journal of Differential Equations, 2006 (138), 1--9, 2006.

        \bibitem{firstdeep}
G. Hinton, S. Osindero, Y-W. Teh: \emph{A fast learning algorithm for deep belief nets},
Neural Computation, 18(7), 1527--1554, 2006.




\bibitem{Hornik1990551}
K. Hornik, M. Stinchcombe, H. White: \emph{Universal approximation of an unknown mapping
and its derivatives using multilayer feedforward networks}m Neural Networks, 3(5), 551--560, 1990.




\bibitem{SciPy} E.~Jones, T.~Oliphant, P.~Peterson, et al., \emph{SciPy: Open source scientific tools for Python}, http://www.scipy.org, 2001–.

\bibitem{KaWa} A.~Karakhanyan, X.~J.~Wang: \emph{On the reflector shape design}, J. Differential Geometry, 84, 561-610, 2010.





\bibitem{LaP}
O. Lakkis, T. Pryer: \emph{A finite element method for nonlinear elliptic problems}, SIAM Journal
on Scientific Computing, 35(4), 2025--2045, 2013.



\bibitem{li-liu:2017-1}
Q.~Li,  Z.Y. Liu: \emph{Solving the 2-{D} elliptic {M}onge-{A}mp{\`e}re equation by a
  {K}ansa's method}, Acta Mathematicae Applicatae Sinica, English Series,
  33(2), 269--276, 2017.


\bibitem{liu-et-al:2017-1}
J.~Liu, B.D. Froese, A.M. Oberman, M.Q. Xiao: \emph{A multigrid scheme for 3{D} {M}onge-{A}mp{\`e}re equations}, {International Journal of Computer Mathematics}, 94(9), 1850--1866, 2017.


\bibitem{Liuu} X. Liu: \emph{Kolmogorov Superposition Theorem and Its
Applications}, PhD thesis, Imperial College of London, 2015.


\bibitem{liu-he:2013-1}
Z.Y. Liu, Y.~He: \emph{Cascadic meshfree method for the elliptic {M}onge-{A}mp{\`e}re equation}, {Engineering Analysis with Boundary Elements}, 37(7), 990--996,
  2013.





\bibitem{Lo} G. G. Lorentz: \emph{Approximation of Functions},  AMS Chelsea Publishing Series, AMS
Chelsea, 2005.






\bibitem{maiorov1999lower}
V. Maiorov,  A. Pinkus: \emph{Lower bounds for approximation by mlp neural networks},
Neurocomputing, 25(1), 81--91, 1999.



\bibitem{mhaskar1996neural}
H. N. Mhaskar: \emph{Neural networks for optimal approximation of smooth and analytic functions},
Neural computation, 8(1), 164--177, 1996.

\bibitem{Mo} C. Mooeny: \emph{The Monge-Amp\`ere Equation}, arXiv:1806.09945, 2018.

 \bibitem{Ne}
  M. Neilan: \emph{Finite element methods for fully nonlinear second order PDEs based on a discrete
Hessian with applications to the Monge-Amp\`ere equation}, Journal of Computational and Applied
Mathematics, 263, 351--369, 2014.

\bibitem{oberman:2008-1}
A.~Oberman: \emph{Wide stencil finite difference schemes for the elliptic
  {M}onge-{A}mp\`ere equations and functions of the eigenvalues of the
  {H}essian}, Discrete Contin. Dyn. Syst. Ser B, 10(1), 221--238, 2008.

\bibitem{Po2}A.~V.~Pogorelov: \emph{The Dirichlet problem for the multidimensional analogue of the Monge-Amp\`re equation}. Dokl. Akad. Nauk SSSR, 201, 790-–793, 1971.


\bibitem{Po}A.~V.~Pogorelov: \emph{On the improper convex affine hyperspheres}, Geometriae Dedicata, 1 (1) 33–46, 1972.

\bibitem{PriBeIjTu} C.~R.~Prins, R.~Beltman, J.~.H.~.M.~Then Thije Boonkkamp, W.~L.~Ijzerman, T.~W.~Tukker: \emph{A least-squares method for optimal transport using the Monge-Amp\`ere equation}, SIAM Journal on Scientific Computing 37(6), 2015.

\bibitem{RaTa} Rauch, Taylor: \emph{The Dirichlet Problem for the Multidimensional Monge-Amp\`ere Equation}, Rocky Mountain Journal of Mathematics, 7(2), 345--364, 1977.

\bibitem{Re} F.~Rellich: \emph{Zur ersten Randwertaufgabe bei Monge-Amporeschen Differentialgleichungen vom elliptischen Typus; differentialgeometrische Anwendungen}, Math. Ann., 107, 505--513, 1933.

    \bibitem{backprop}
D. Rumelhart, G. Hinton, R. Williams: \emph{Learning representations by back-propagating
errors}, Nature, 323(6088), 533--536, 1986.





\bibitem{dropout2}
N. Srivastava, G. Hinton, A. Krizhevsky, I. Sutskever, R. Salakhutdinov: \emph{Dropout: A
simple way to prevent neural networks from overfitting}, Journal of Machine Learning
Research, 15, 1929--1958, 2014.

\bibitem{TruWa2} N.~Trudinger, X.~J.~Wang: \emph{The Monge–Ampère equation and its applications}, Handbook of geometric analysis. No. 1, 467–524, Adv. Lect. Math. (ALM), 7, Int. Press, Somerville, MA, 2008.

\bibitem{TruWa} N.~Trudinger and X.~J.~Wang: \emph{On the second boundary value problem for Monge-Amp\`ere type equations and optimal transportation}, Ann. Scuola Norm. Sup. Pisa Cl. Sci. (5), Vol. VIII, 143-174, 2009.

\bibitem{Ur} J.~Urbas: \emph{On the second boundary value problem for equations of Monge-Amp\`ere type}, J. reine. angew. Math., 487, 115-124, 1997.



\bibitem{we1}
E. Weinan: \emph{Machine learning and computational mathematics},  ArXiv e-prints,
arXiv:2009.14596, 2020.

\bibitem{we2}
E. Weinan, S. Wojtowytsch: \emph{On the banach spaces associated with multi-layer relu networks:
Function representation, approximation theory and gradient descent dynamics},
ArXiv e-prints, arXiv: 2007.15623, 2020.

\bibitem{we3}
E. Weinan, C. Ma, S. Wojtowytsch, L. Wu: \emph{Towards a mathematical understanding
of neural network-based machine learning: what we know and what we don't}, ArXiv
e-prints, arXiv: 2009.10713, 2020.


  \bibitem{Well}
  H.Weller, P. Browne, C. Budd, M. Cullen, M: \emph{Mesh adaptation on the sphere using optimal
transport and the numerical solution of a Monge-Amp\`ere type equation}, Journal of Computational
Physics, 308, 102--123, 2016



\bibitem{2018arXiv181208883X}
K. Xu, E. Darve: \emph{Calibrating L\'evy process from observations based on neural
networks and automatic differentiation with convergence proofs},  ArXiv e-prints, arXiv:1812.08883v1, 2018.


\end{thebibliography}
\end{document}